\pdfoutput=1

\documentclass[11pt]{article}

\usepackage[preprint]{acl}

\usepackage{times}
\usepackage{latexsym}

\usepackage[T1]{fontenc}

\usepackage[utf8]{inputenc}

\usepackage{microtype}

\usepackage{inconsolata}

\usepackage{graphicx}
\usepackage{multirow}
\usepackage{wrapfig}
\usepackage{amsmath}
\usepackage{amssymb}
\usepackage{caption}
\usepackage{subcaption}
\usepackage[inline]{enumitem}
\usepackage{xspace}
\usepackage{booktabs}
\usepackage{wrapfig}
\usepackage{colortbl}
\usepackage{tikz}
\usepackage{tabularx}
\usepackage{comment}
\usepackage{hyperref}
\usepackage{url}

\def\BenchmarkName{\texttt{VisualWebArena}\xspace}

\title{\texttt{VisualWebArena}: Evaluating Multimodal Agents on Realistic Visually Grounded Web Tasks}

\author{%
Jing Yu Koh \quad 
Robert Lo\thanks{Equal contribution.} \quad
Lawrence Jang\footnotemark[1]  \quad Vikram Duvvur\footnotemark[1] \\
\textbf{Ming Chong Lim}\footnotemark[1]  \quad \textbf{Po-Yu Huang}\footnotemark[1]  \quad \textbf{Graham Neubig} \quad \textbf{Shuyan Zhou}  \\ 
\textbf{Ruslan Salakhutdinov} \quad \textbf{Daniel Fried}\\ 
{Carnegie Mellon University} \\
  \texttt{\{jingyuk,rsalakhu,dfried\}@cs.cmu.edu} \\
}

\begin{document}
\maketitle
\begin{abstract}
Autonomous agents capable of planning, reasoning, and executing actions on the web offer a promising avenue for automating computer tasks. However, the majority of existing benchmarks primarily focus on text-based agents, neglecting many natural tasks that require visual information to effectively solve. Given that most computer interfaces cater to human perception, visual information often augments textual data in ways that text-only models struggle to harness effectively.
To bridge this gap, we introduce \BenchmarkName{}, a benchmark designed to assess the performance of multimodal agents on realistic \textit{visually grounded web tasks}. \BenchmarkName{} comprises of diverse and complex web-based tasks that evaluate various capabilities of autonomous multimodal agents. 
To perform well, agents need to accurately process image-text inputs, interpret natural language instructions, and execute actions on websites to accomplish user-defined objectives.
We evaluate state-of-the-art LLM-based autonomous agents, including several multimodal agents. Our analysis reveals several limitations of text-based LLM agents, gaps in the capabilities of state-of-the-art multimodal language agents, and insights towards building stronger autonomous agents for the web. 
\end{abstract}

\section{Introduction}
Automating routine computer tasks with autonomous agents is a long standing goal of artificial intelligence research~\citep{franklin1996agent,jennings1998roadmap}. To achieve this, we need agents that can navigate computers effectively, process visual and textual inputs, handle high-level natural language instructions, and execute actions to achieve desired goals. As digital interfaces today are primarily built for human eyes, effective visual understanding is necessary for many routine computer tasks. For example, humans frequently perform tasks on the web which involve visual references, such as ``Help me order a green polo shirt from Amazon,'' or rely on pictures rather than text to communicate. 
However, many agent benchmarks today focus on text-based tasks, neglecting the evaluation (and consequently the development) of multimodal agents.
\begin{figure*}[t]
\centering
\includegraphics[width=1.0\linewidth]{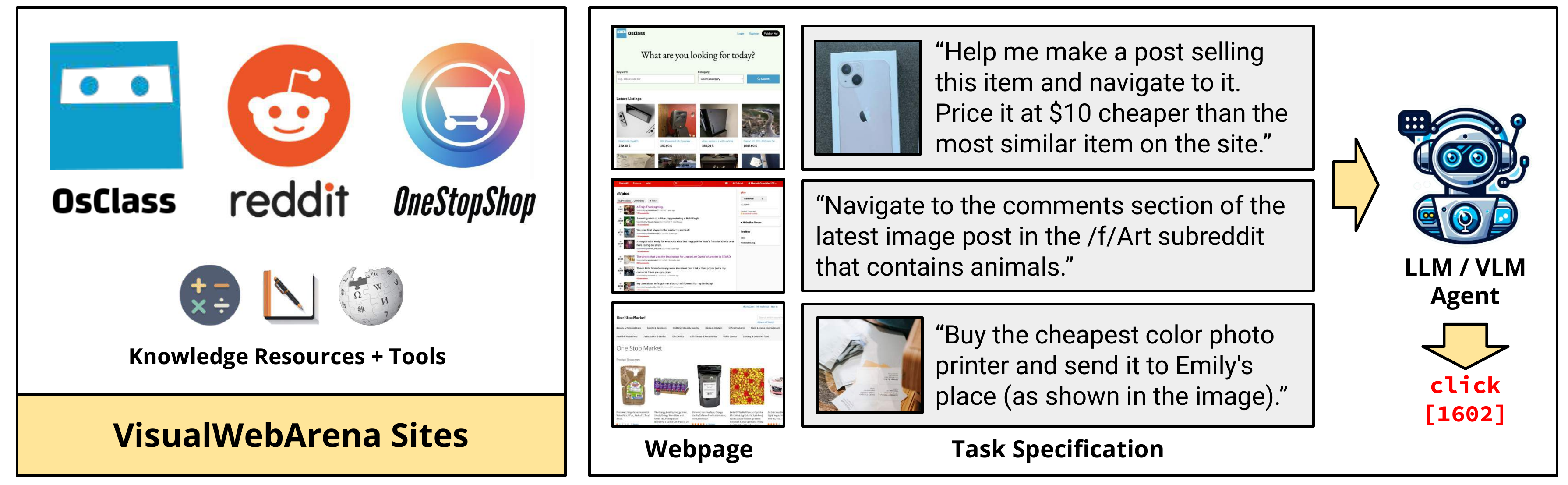}
\caption{\BenchmarkName{} is a benchmark suite of 910 realistic, visually grounded tasks on self-hosted web environments that involve web navigation and visual understanding.
}
\label{fig:task_overview}
\end{figure*}
To address this gap, we propose \BenchmarkName{}~(Fig.~\ref{fig:task_overview}), a benchmark suite designed to rigorously assess and advance the visual and textual capabilities of autonomous agents. 
\BenchmarkName{} builds off the WebArena~\citep{zhou2023webarena} framework, leveraging reproducible self-hosted environments and execution-based evaluations. 
\BenchmarkName{} introduces a set of unique tasks that emphasize integrating visual understanding with language processing, closely simulating human interaction with modern computing interfaces. 
Our contributions are summarized as follows:

\begin{itemize}
    \item We introduce \BenchmarkName{}, a set of 910 realistic tasks over three diverse web environments: Classifieds, Shopping, and Reddit. The Classifieds environment is a new contribution with real world data, while the Shopping and Reddit environments are inherited from WebArena. All tasks we introduce are \textit{visually grounded}, and require visual understanding of webpage content to effectively solve (while WebArena does not). 25.2\% of the tasks also include images as input (Fig.~\ref{fig:task_overview}), and require understanding interleaved image-text inputs. 
    \item We extensively benchmark the autonomous capabilities of state-of-the-art (SOTA) large language models (LLM) and vision-language models (VLMs), demonstrating that strong VLMs outperform text-based LLMs. The best VLM agents achieve a success rate of 16.4\% on \BenchmarkName{}, which is still significantly below human performance of 88.7\%. 
    \item We propose a new VLM agent inspired by Set-of-Marks prompting~\cite{yang2023som}, simplifying the action space of the model. We show that this model substantially outperforms other baseline LLM agents, especially on sites that are more visually complex.
\end{itemize}

\section{Related Work}

\paragraph{Language-Guided Web Agent Benchmarks}
The development of reproducible environments for autonomous agents has seen considerable progress in recent years. Earlier efforts introduced reinforcement learning environments~\citep{brockman2016openai}, and extended into web domains~\citep{shi2017world,liu2018reinforcement}. 
Recent web agent benchmarks introduced tasks involving actions on static internet pages~\citep{deng2023mind2web} as well as interaction in simulated web environments~\citep{yao2022webshop, zhou2023webarena}. AgentBench~\citep{liu2023agentbench} extends the scope of agents for computer interaction beyond the web, exploring database management and operating system functionalities.

\paragraph{LLM Agents}
There has been significant recent interest in using Large Language Models (LLMs) for developing autonomous agents~\citep{xi2023rise,wang2023survey}. State-of-the-art LLMs~\citep{team2023gemini,openai2023gpt4,chowdhery2023palm,rae2021gopher,zhang2022opt,touvron2023llama,touvron2023llama2,jiang2023mistral,jiang2024mixtral} based on the Transformer~\citep{vaswani2017attention} architecture have demonstrated impressive abilities in learning from in-context examples~\citep{brown2020language,chan2022transformers}, reasoning~\citep{wei2022chain,yao2023tree,wang2022self,besta2023graph}, following instructions~\citep{chung2022scaling,longpre2023flan,ouyang2022training}, and operating over long-context sequences~\citep{tay2020long,bertsch2023unlimiformer,tworkowski2023focused}. Several recent works leverage these abilities for building autonomous web agents: \citet{kim2023language} propose a recursive prompting method to improve GPT-4 performance on MiniWoB++~\citep{liu2018reinforcement}. \citet{liu2023bolaa} propose a method of orchestrating multiple LLM agents to improve performance on the WebShop~\citep{yao2022webshop} environment. \citet{zeng2023agenttuning} fine-tunes the LLaMA-2 models on interaction trajectories with instructions, improving over baseline agents.

\paragraph{Vision-Language Models}
Finally, our work builds off advances in vision-language models (VLMs), used for many multimodal tasks such as image captioning~\citep{vinyals2015show}, visual question answering~\citep{antol2015vqa}, and other benchmarks~\citep{mialon2023gaia,yue2023mmmu,tong2024eyes}.
Frozen~\citep{tsimpoukelli2021multimodal} was one of the first approaches to demonstrate the effectiveness of finetuning a visual encoder to map images into the embedding space of a LLM, introducing compelling few-shot multimodal abilities. \citet{alayrac2022flamingo} introduced cross-attention layers and scaled up model sizes and training data. \citet{wang2023cogvlm} introduced trainable visual expert modules to improve vision-language fusion. \citet{liu2023visual} proposed fine-tuning on images paired with instructions to improve text generation performance on several multimodal tasks. GPT-4V~\citep{openai2023gpt4} introduces visual processing to the GPT-4 family of models. 
Gemini~\citep{team2023gemini} is multimodal from the beginning (in contrast to post-hoc fine-tuned models), and can handle text interleaved with visual and audio inputs. 
Several recent work have also explored using VLMs to build agents for mobile platforms~\citep{zhan2023you,chu2023mobilevlm,yang2023appagent} and the web~\citep{gur2023real,hong2023cogagent}. \citet{zheng2024gpt} is contemporaneous work which performs action grounding to identify appropriate HTML elements for enabling agents to execute actions. In contrast, our proposed SoM agent uses JavaScript to produce a Set-of-Marks~\citep{yang2023som} for the VLM to directly use as an observation and action space.

\section{\BenchmarkName{} Environment} 

In order to ensure reproducibility, realism, and determinism, all websites in the \BenchmarkName{} framework are provided as standalone self-hosted web applications. The textual and visual content are acquired from real world counterparts, while the code is based off open-source infrastructure commonly used in real websites. We formally define the environment, observation space, and action space below, but encourage readers to refer to WebArena~\citep{zhou2023webarena} for more details.

The environment and agent can be modeled as a partially observable Markov decision process (POMDP): $\mathcal{E} = (S, A, \Omega, T)$, where $S$ represents the set of states, $A$ represents the set of actions (Sec.~\ref{sec:action_space}), and $\Omega$ represents the set of observations (Sec.~\ref{sec:observation_space}). The transition function is defined as $T: S \times A \rightarrow S$, with deterministic transitions between states conditioned on actions. At each time step $t$, the environment is in some state $s_t$ (e.g., a particular page), with a partial observation $o_t \in \Omega$. An agent issues an action $a_t \in A$ conditioned on $o_t$, which results in a new state $s_{t+1} \in S$ and a new partial observation $o_{t+1} \in \Omega$ of the resulting page. The action $a_t$ may be an action to be executed on the webpage (Tab.~\ref{tab:actions}), or it may simply be a string output for information seeking tasks (Sec.~\ref{sec:environment_evaluation}).

Finally, we define the reward function $R: S \times A \rightarrow \{0, 1\}$ (Sec.~\ref{sec:environment_evaluation}) to measure the success of a task execution. In \BenchmarkName{}, the reward function returns 1 at the final step if the state transitions align with the expectations of the task objective (i.e., the goal is achieved), and 0 otherwise. 

\subsection{Observation Space} \label{sec:observation_space}

\begin{figure*}[t]
    \centering
    \includegraphics[width=1.0\linewidth]{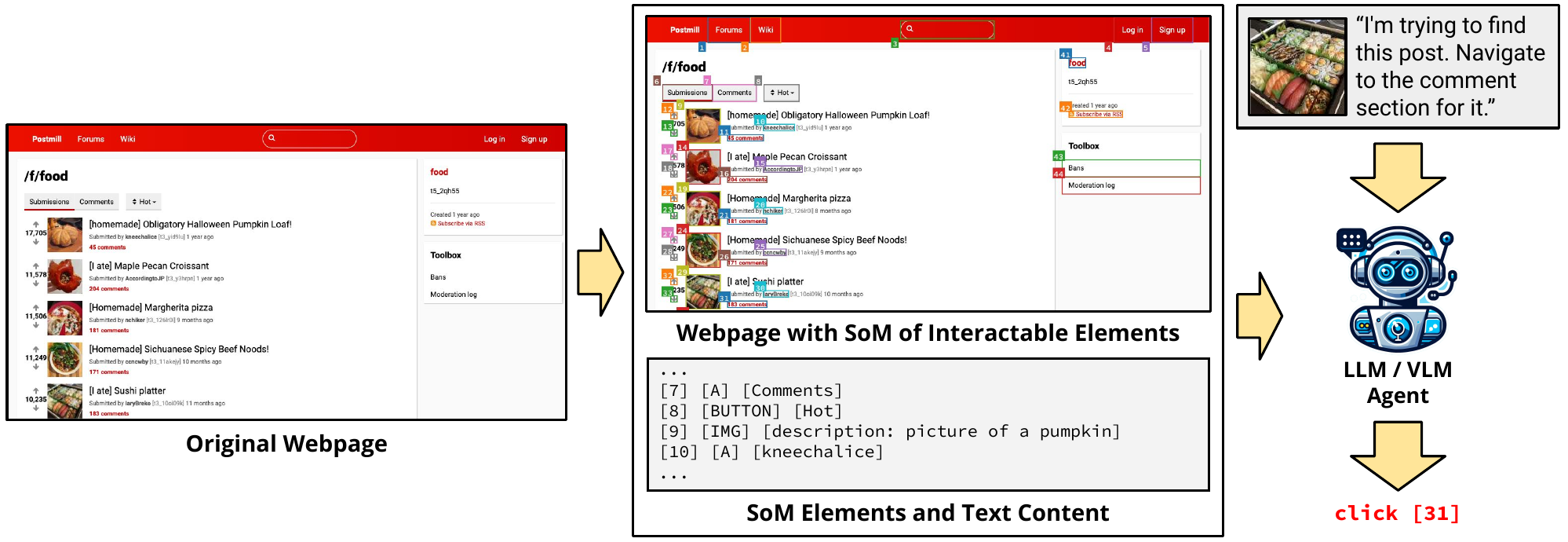}
    \caption{Set-of-Marks~\citep{yang2023som} augmented webpage screenshot. Every interactable element is highlighted with a bounding box and a unique ID.}
    \label{fig:som_figure}
\end{figure*}

The observation space $\Omega$ is modeled after a realistic web browsing experience. Observations include the webpage URLs, opened tabs (possibly multiple tabs of different websites), and the webpage content of the focused tab. In 25.2\% of tasks, the intent also involves one or more input images (e.g., the first and third tasks in Fig.~\ref{fig:task_overview}). The webpage content can be represented in several ways:
\begin{enumerate}
\item Raw web page HTML as a Document Object Model (DOM) tree, used in previous works on autonomous web agents~\citep{shi2017world,liu2018reinforcement,deng2023mind2web}.
\item The accessibility tree,\footnote{\url{https://developer.mozilla.org/en-US/docs/Glossary/Accessibility_tree}} which provides a structured and simplified representation of the webpage content that is optimized for assistive technologies. This is the primary representation that WebArena~\citep{zhou2023webarena} uses for its baseline LLM agents.
\item Web screenshots as RGB arrays, which has demonstrated efficacy in prior work~\citep{gur2023real,hong2023cogagent,yan2023gpt}.
\item We introduce a new visual representation inspired by Set-of-Marks (SoM) prompting~\citep{yang2023som}. For every interactable element on the webpage, we label it with a bounding box and an ID (Fig.~\ref{fig:som_figure}), producing a screenshot for visual agents to reference elements on the page using their unique ID. We provide more details and analysis in Sec.~\ref{sec:som_method}.
\end{enumerate}

\subsection{Action Space} \label{sec:action_space}
\begin{table} 
\vspace{-0.2in}
\centering
\resizebox{1.0\linewidth}{!}{%
\begin{tabular}{ll}
  \toprule
  \textbf{Action Type $a$}\hspace{25mm} & \textbf{Description} \\
  \midrule
  \texttt{click [elem]} & Click on element \texttt{elem}. \\
  \texttt{hover [elem]} & Hover on element \texttt{elem}. \\
  \texttt{type [elem] [text]} & Type \texttt{text} on element \texttt{elem}. \\
  \texttt{press [key\_comb]} & Press a key combination. \\
  \texttt{new\_tab} & Open a new tab. \\
  \texttt{tab\_focus [index]} & Focus on the i-th tab. \\
  \texttt{tab\_close} & Close current tab. \\
  \texttt{goto [url]} & Open \texttt{url}. \\
  \texttt{go\_back} & Click the back button. \\
  \texttt{go\_forward} & Click the forward button. \\
  \texttt{scroll [up|down]} & Scroll up or down the page. \\
  \texttt{stop [answer]} & End the task with an output. \\
  \bottomrule
\end{tabular}
}
\caption{Set of possible actions $A$.}   \label{tab:actions}
\end{table}

The full set of actions $A$ is summarized in Tab.~\ref{tab:actions}. 
The arguments for action $a_t$ is the unique element ID from the current observation $o_t$. 
An advantage of this representation (over predicting $(x, y)$ coordinates) is that it allows us to focus on high level reasoning rather than low-level control, as many SOTA VLMs and LLMs were not explicitly trained for referencing elements at such fine granularity.
For the agents with accessibility tree representations, the argument is the element ID in the tree. For the SoM representation, we use the unique IDs assigned in the current page (see Fig.~\ref{fig:som_figure}).

\subsection{Evaluation}   \label{sec:environment_evaluation}

In order to evaluate performance on \BenchmarkName{}, we introduce new visually grounded evaluation metrics to the functional evaluation paradigm of WebArena. These allow us to comprehensively evaluate the correctness of execution traces on open ended visually grounded tasks. The rewards for each task are hand designed functions using the primitives described below.

\paragraph{Information Seeking Tasks}
Information seeking tasks (e.g., the first task in Tab.~\ref{tab:evaluation}) expect a string output $\hat{a}$ from the model. We adopt similar reward functions as WebArena for measuring text correctness against a groundtruth output $a^*$:
\begin{itemize}
    \item \texttt{exact\_match}: This can be defined as $\mathbf{1}_{\{\hat{a} = a^*\}}$. 
    Only outputs that are exactly equal to the groundtruth are given a score of 1. This is used in tasks where an exact response (e.g., a numerical answer) is expected.
    \item \texttt{must\_include}: This reward function gives a score of 1 if all elements in $a^*$ are contained in $\hat{a}$ and 0 otherwise. For example, if $\hat{a} = \text{``\$1.99, \$2.50, \$10.00''}$ and $a^* = \{\text{``1.99''}, \text{``2.50''}, \text{``10.00''} \}$, the task is awarded a score of 1 as all expected elements are present in the output. This is primarily used in tasks where we expect an unordered list of outputs, or we expect text output to contain a particular keyword.
    \item \texttt{fuzzy\_match}: This function queries a LLM (GPT-4-Turbo in our implementation) to evaluate whether $a^*$ and $\hat{a}$ are semantically equal. The LLM is prompted to output ``correct'', ``incorrect'', or ``partially correct'', and we assign a reward of 1 if the output is ``correct''.\footnote{We do not consider non-binary rewards in this work, but this would be a valuable direction to explore in the future towards introducing more continuous scales of performance.} This evaluation is useful for more open ended settings where we are only concerned with semantic rather than exact equivalence, such as asking the user to add a comment describing an image.
    \item \texttt{must\_exclude}: We introduce this function, which is the converse of \texttt{must\_include}. A reward of 0 is assigned if any element from a set $a^*$ is found in $\hat{a}$ (and 1 otherwise). For instance, if $\hat{a} = \text{``\$1.99, \$2.50, \$10.00''}$ and $a^* = \{\text{``1.50''}, \text{``2.00''}\}$, the reward is 1 as none of the prohibited elements are in the output.
\end{itemize}

In addition, we also introduce several new visual functions for measuring open ended tasks:
\begin{itemize}
    \item \texttt{eval\_vqa}: Similar to \texttt{fuzzy\_match}, this function queries a VLM capable of performing visual question answering (VQA)~\citep{antol2015vqa}. We use BLIP-2-T5XL~\citep{li2023blip} in our implementation. We query the VLM with an image and a question. If the output of the VLM contains the groundtruth answer $a^*$, a reward of 1 is assigned. 
    This is useful for evaluating more open ended tasks, e.g., ``Buy me a green hoodie under \$10.''. There are many possible products that satisfy this objective, and it would be infeasible to enumerate all their IDs.
    \item \texttt{eval\_fuzzy\_image\_match}: This function checks whether a query image is similar to a groundtruth image according to the structural similarity index measure (SSIM)~\citep{wang2004image}. If the SSIM between the query and groundtruth images is higher than a threshold $t \in [0, 1]$, a reward of 1 is assigned. 
\end{itemize}

\paragraph{Navigation and Actions} 

Many tasks in \BenchmarkName{} require navigating through multiple webpages, and executing actions to change the underlying state $s$ of the environment. To accurately evaluate certain objectives, we require reward functions that examine the final webpage state to determine whether the task was successfully accomplished. 
Each evaluator consists of a \texttt{locator} as well as a URL. The URL can be a specific page, or a function (e.g., the last page that the agent navigated to). The \texttt{locator} describes the object on the page that should be examined (e.g., all \texttt{img} elements, or all elements with the \texttt{.product-image-photo} class). During evaluation, we use the locator to retrieve the corresponding image or text content, and reuse the functions from the information seeking tasks to check for correctness.

\begin{table*}[t]
\resizebox{\linewidth}{!}{%
\begin{tabular}{lp{.35\linewidth}l}
  \toprule
  \textbf{Webpage / Input Image(s)} &   \textbf{Example Intent}  & \textbf{Reward Function $r(s, a)$ Implementation}  \\
  \midrule
  \multirow{2}{*}{\raisebox{-2cm}{\includegraphics[width=3.5cm]{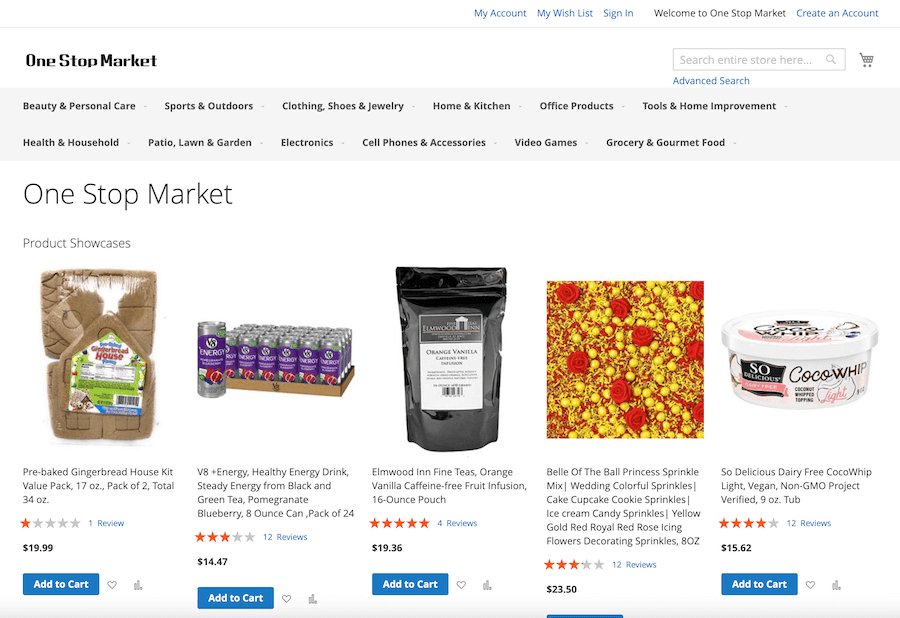}}}  &  \multirow{2}{*}{\begin{minipage}{.35\textwidth}Buy the least expensive red blanket from the ``Blankets \& Throws'' category.\end{minipage}} &  \texttt{url="func:shopping\_get\_latest\_order\_url"}  \\
    &    &  \texttt{must\_include($\hat{a}$, \{ "B0983XCYK6", "Red" \})}  \\
     &   &    \\ &   &    \\  &   &    \\ 
  \midrule
  \multirow{2}{*}{\raisebox{-1cm}{\includegraphics[width=2cm,height=2cm]{figures/images/hotdog_man.jpg}}} &  \multirow{2}{*}{
   \begin{minipage}{1.0\linewidth}
   Add something like what the man is wearing to my wish list.
   \end{minipage}} &  \texttt{url="/wishlist"} \\
   &   &  \texttt{locator(".wishlist .product-image-photo")}  \\
   &   &  \texttt{eval\_vqa($s$, "Is this a polo shirt? (yes/no)", "yes")}   \\ 
   &   & \texttt{eval\_vqa($s$, "Is this shirt green? (yes/no)", "yes")}   \\
  \midrule
  \raisebox{-1.2cm}{\includegraphics[width=1.5cm,height=1.5cm]{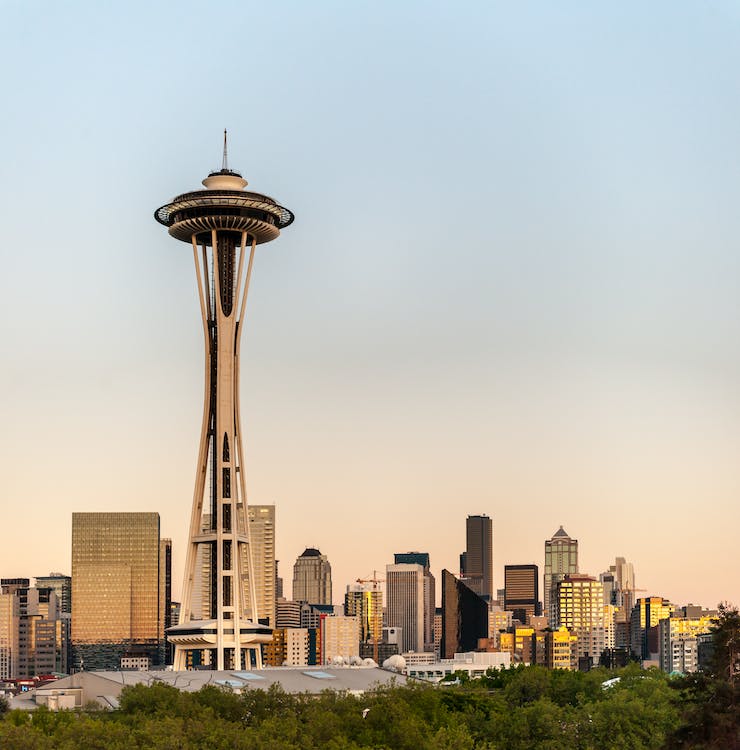}} \raisebox{-1.2cm}{\includegraphics[width=1.5cm,height=1.5cm]{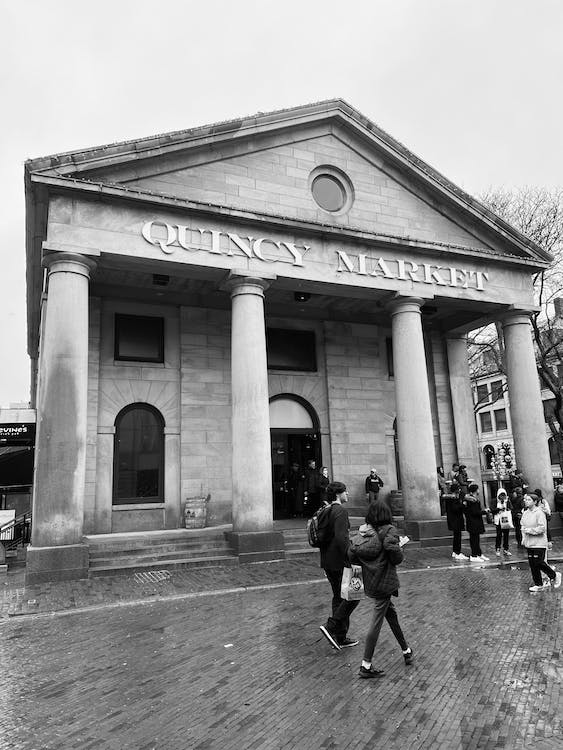}}  &  Create a post for each of these images in the most related forums. &  \texttt{eval\_fuzzy\_image\_match($s$, $a^*$)}   \\
  \midrule
   \multirow{4}{*}{\includegraphics[width=3.5cm]{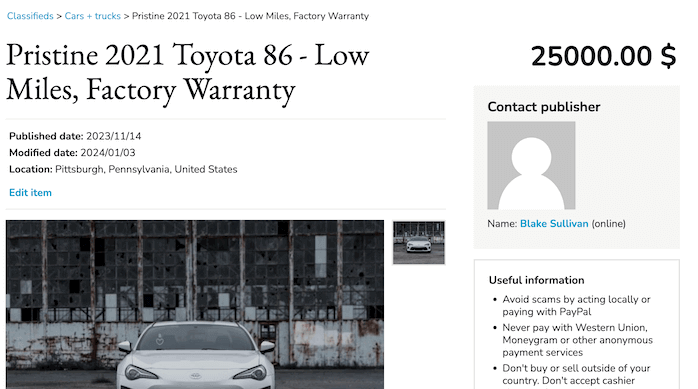}} &  \multirow{4}{*}{
   \begin{minipage}{1.0\linewidth}
   Navigate to my listing of the white car and change the price to \$25000. Update the price in the description as well.
   \end{minipage}} &  \texttt{url="/index.php?page=item\&id=84144"} \\
   & & \texttt{must\_include($\hat{a}$, "\$25000 |OR| \$25,000")} \\
   & & \texttt{must\_exclude($\hat{a}$, "\$30000 |OR| \$30,000")} \\
   & &  \\
  \bottomrule
\end{tabular}
}
\caption{Various evaluation metrics to assign reward $r(s, a) \in R: S \times A \rightarrow \{0, 1\}$. Our execution-based reward primitives allow us to benchmark many diverse, realistic, and open-ended tasks.} \label{tab:evaluation}
\end{table*}

\section{Curating Visually Grounded Tasks}

\subsection{Web Environments}
\BenchmarkName{} is designed around three realistic web environments that involve visually rich content. Several tasks require referencing information from a self-hosted Wikipedia knowledge base, and others involve interacting across more than one website.

\paragraph{Classifieds}
We introduce a new Classifieds website in \BenchmarkName{}, inspired by real world marketplaces such as Craigslist and Facebook Marketplace. The Classifieds site contains 65,955 listings and provides a distinct environment compared to existing ones in WebArena, introducing visually grounded tasks centered around user interactions typical in classifieds websites (posting, searching, commenting). The site's infrastructure uses \href{https://osclass-classifieds.com}{OSClass}, a robust open-source Content Management System (CMS) designed for classifieds ads, used in multiple real world sites. OSClass enables functions such as search, posting, commenting, and leaving reviews and ratings. More details about the environment are provided in Appendix.~\ref{sec:appendix_classifieds}. 

\paragraph{Shopping}

The Shopping site follows the e-commerce environment from WebArena~\citep{zhou2023webarena}, with product information and content scraped from Amazon and released in WebShop~\citep{yao2022webshop}. Visual understanding of product images is required for successfully navigating and completing tasks on e-commerce platforms, making this a natural choice for \BenchmarkName{}. 

\paragraph{Reddit}

The Reddit site also follows the same environment from WebArena, and represents a social forum platform. The site contains 31,464 posts containing a diverse set of images across different subreddits and forums, such as natural images, memes, consumer electronics, and charts. 

\subsection{Tasks} \label{sec:tasks}



\paragraph{Task Creation} We introduce a set of 910 new tasks, split across the three sites detailed earlier. 
We focus on curating realistic visually grounded tasks, following a similar process as task creation in WebArena. We start by having 6 graduate students (co-authors of this paper) write \textit{intent templates} (e.g., ``Find me the \texttt{\{\{attribute\}\}} \texttt{\{\{item\}\}}. It should be between \texttt{\{\{range\}\}}.''), which can be manually expanded by the annotator to form multiple tasks (e.g., ``Find me the cheapest red Toyota. It should be between \$3000 to \$6000.''). 
We encouraged the annotators to be creative, and make use of the visual layouts of the websites, input images, and cross-site functionalities to develop creative and realistic tasks. When tasks include input images, these were sourced from royalty-free, attribution-free sources and MS-COCO~\citep{lin2014microsoft}. Annotators also wrote the reward functions using the primitives described in Sec.~\ref{sec:environment_evaluation}. We collected a total of 314 unique templates (average of 2.9 tasks per template). 
While the majority of tasks can be solved, we also included a small subset (46 tasks, or 5.1\%) which are unachievable. This subset tests the ability of agents to terminate early in the event where a task cannot be solved, which is essential in many real world scenarios. For unachievable tasks, we require agents to output a reason why the task is unachievable, which is evaluated using the \texttt{fuzzy\_match} function (Sec.~\ref{sec:environment_evaluation}).

\paragraph{Visually Grounded Tasks} A key aspect of \BenchmarkName{} is the inherent visual grounding of all tasks. Each task demands visual understanding, requiring agents to process and interpret visual information rather than relying solely on textual or HTML-based cues. This aligns closely with modern human-computer interfaces, where visual information (e.g., icons, colors) is often critical. For instance, a typical task might involve selecting a visually specific item, such as a ``green polo shirt'' where the color is visually discernible but not explicitly mentioned in text.

\paragraph{Task Complexity}

We classify each task into three difficulty levels: \textbf{easy}, \textbf{medium}, and \textbf{hard}. This classification is particularly useful for assessing performance across a spectrum of agents, ranging from smaller models to state-of-the-art LLMs and VLMs. We find in our analysis (Sec.~\ref{sec:baselines}) that many open-source models (e.g., LLaMA-2-70B, IDEFICS-80B) achieve a success rate of close to 0 on medium or hard tasks, but non-zero performance on easy tasks. This suggests that running open-source models on the easy subset would provide useful signal during development as well as faster iteration cycles (assuming performance between weaker and stronger agents are correlated). We also provide more detailed analysis in Appendix.~\ref{sec:performance_difficulty}.

We annotate both the \textit{action} and \textit{visual} difficulty of each task. 
The action difficulty is determined by the estimated number of actions that a human would need to complete the task. \textbf{Easy} tasks are defined as those that require three or fewer actions, \textbf{medium} tasks involve four to nine actions, and \textbf{hard} tasks demand ten or more. 
Visual difficulty is similarly segmented: 
\textbf{Easy} tasks involve basic visual identification of colors, shapes, and high-level object detection (e.g., recognizing the presence of a cat). 
\textbf{Medium} tasks require discerning patterns, semantic understanding, or OCR on large text of shorter lengths. 
\textbf{Hard} tasks involve multiple images, OCR on small or lengthy text, or fine details. 
Finally, the overall difficulty level is determined by averaging the visual and reasoning complexities. However, human judgment may lead to deviations in this assessment, as certain tasks might inherently skew more towards primarily testing visual or reasoning challenges.

\subsection{Human Performance}

We measure the success rate of 7 college students on \BenchmarkName{} tasks. Several of these students also assisted with task creation, and to avoid data leakage, we ensured that they were not assigned to the same tasks that they initially created. We sample one task per template, collecting a representative set of 230 tasks. We find that humans do well at these tasks, achieving an overall success rate of 88.7\% (Tab.~\ref{tab:results}). The mistakes made in the remaining tasks are usually minor, such as not reading the task correctly or missing a part of the objective. For example, one task asked to add a particular item to the wishlist, but the human added it to the shopping cart instead. 
Another common failure mode was for tasks that required exhaustive search (e.g., ``Find and navigate to the comments of this exact image.''). Users were often unable to find the appropriate post after searching for 5--10 minutes and gave up, assuming that the task was unachievable. In many shopping tasks, humans also did not look through all possible candidate pages to identify the cheapest or most highly reviewed product. We found these failure modes interesting, as they represent issues that strong agents would be well poised to handle, potentially achieving above human performance and speed.
\section{Baselines}  \label{sec:baselines}

We run several baselines to benchmark the performance of state-of-the-art LLM and VLM agents. All models are prompt-based and provided with 3 in-context examples (one from each environment), which share no overlap with the benchmark tasks. The prompts we use are provided in the appendix. We summarize the results in Tab.~\ref{tab:results} and describe the baselines in detail in the following sections.

\subsection{Text-based LLM Agents}

Several prior works developed autonomous agents by prompting text-based LLMs~\citep{zhou2023webarena,kim2023language,liu2023bolaa}. We benchmark several text-based LLM agents with Chain-of-Thought prompting~\cite{wei2022chain} over the accessibility tree representations of the websites as input, and leave more advanced prompting strategies for future work. We test API-based LLMs, including GPT-4 Turbo (\texttt{gpt-4-1106-preview}), GPT-3.5 Turbo (\texttt{gpt-3.5-turbo-1106}), and Gemini-Pro, as well as open sourced LLMs (LLaMA-2-70B, Mixtral-8x7B).

\begin{table*}[t]
\centering
\resizebox{\linewidth}{!}{%
\begin{tabular}{@{}lccccccc@{}}
\toprule
\multirow{2}{*}{\textbf{Model Type}} & \multirow{2}{*}{\textbf{LLM Backbone}} & \multirow{2}{*}{\textbf{Visual Backbone}} & \multirow{2}{*}{\textbf{Inputs}} & \multicolumn{4}{c}{\textbf{Success Rate ($\uparrow$)}} \\
\cmidrule(lr){5-8}
& & & & \textbf{Classifieds} & \textbf{Reddit} & \textbf{Shopping} & \textbf{Overall} \\
\midrule
\multirow{5}{*}{Text-only} & LLaMA-2-70B & \multirow{5}{*}{-} & \multirow{5}{*}{Acc. Tree} & 0.43\% & 1.43\% & 1.29\% & 1.10\% \\
                        & Mixtral-8x7B &  &  & 1.71\% & 2.86\% & 1.29\% & 1.76\% \\
                        & Gemini-Pro &  &  & 0.85\% & 0.95\% & 3.43\% &  2.20\% \\
                        & GPT-3.5 &  &  & 0.43\% & 0.95\% & 3.65\% &  2.20\% \\
                        & GPT-4 &  &  & 5.56\% & 4.76\% & 9.23\% & 7.25\% \\
\midrule
\multirow{6}{*}{Caption-augmented} & LLaMA-2-70B & BLIP-2-T5XL & \multirow{6}{*}{Acc. Tree + Caps} & 0.00\%  &  0.95\%  & 0.86\% & 0.66\% \\
                        & Mixtral-8x7B & BLIP-2-T5XL &  & 1.28\% & 0.48\% & 2.79\% & 1.87\% \\
                        & GPT-3.5 & LLaVA-7B &  & 1.28\% & 1.43\% & 4.08\% & 2.75\% \\
                        & GPT-3.5 & BLIP-2-T5XL &  & 0.85\%  & 1.43\% & 4.72\% & 2.97\% \\
                        & Gemini-Pro & BLIP-2-T5XL &  & 1.71\%  & 1.43\% & 6.01\% & 3.85\% \\
                        & GPT-4 & BLIP-2-T5XL &   & 8.55\% &  8.57\% &  16.74\% & 12.75\% \\
\midrule
\multirow{4}{*}{Multimodal} & \multicolumn{2}{c}{IDEFICS-80B-Instruct} & \multirow{4}{*}{Image + Caps + Acc. Tree} &  0.43\% & 0.95\% & 0.86\% & 0.77\% \\ 
                        & \multicolumn{2}{c}{CogVLM} &  & 0.00\% & 0.48\% & 0.43\% &  0.33\% \\ 
                        & \multicolumn{2}{c}{Gemini-Pro} &  & 3.42\% & 4.29\% & 8.15\% &  6.04\% \\ 
                        & \multicolumn{2}{c}{GPT-4V} &  & 8.12\%   &  12.38\% &  \textbf{19.74\%} & 15.05\% \\
\midrule
\multirow{4}{*}{Multimodal (SoM)} &  \multicolumn{2}{c}{IDEFICS-80B-Instruct} & \multirow{4}{*}{Image + Caps + SoM} & 0.85\% & 0.95\% & 1.07\% & 0.99\% \\ 
                        & \multicolumn{2}{c}{CogVLM} &  & 0.00\% & 0.48\% & 0.43\% &  0.33\% \\ 
                        & \multicolumn{2}{c}{Gemini-Pro} &  & 3.42\% & 3.81\% & 7.73\% & 5.71\% \\ 
                        & \multicolumn{2}{c}{GPT-4V} &  &  \textbf{9.83\%} &  \textbf{17.14\%} & 19.31\% & \textbf{16.37\%}  \\ 
\midrule
\multirow{1}{*}{Human Performance} &   - & - & \multirow{1}{*}{Webpage} & 91.07\% &  87.10\% & 88.39\% &  88.70\% \\ 
\bottomrule
\end{tabular}
}
\caption{Success rates of baseline LLM and VLM agents on \BenchmarkName{}.}  \label{tab:results}
\end{table*}

\subsection{Image Caption Augmented LLM Agents}

\BenchmarkName{} is a visually grounded benchmark, and we expect that leveraging complementary visual information would improve performance. Hence, we run pretrained image captioning models on every \texttt{img} element on the HTML page, and augment the accessibility tree with this information as the image alt-text before passing this as input to the LLM agents. If a task contains input images, we also caption them and include the captions as part of the prompt. We run experiments on GPT-3.5 with two recent image captioning models, BLIP-2-T5XL~\citep{li2023blip} and LLaVA-v1.5-7B~\citep{liu2023improved}. Our results with GPT-3.5 as the LLM backbone (``Caption-augmented'' section of Tab.~\ref{tab:results}) suggest that the LLaVA and BLIP-2 captioning models achieve comparable performance. Since BLIP-2 achieves a slightly higher success rate, is a smaller model, and requires less GPU VRAM, we use it as the captioning backbone for the remaining experiments.

\subsection{Multimodal Agents} \label{sec:som_method}

Finally, we benchmark strong API-based and open-source VLMs as agents. 
We evaluate several models capable of processing multiple interleaved image-and-text inputs: GPT-4V~\citep{openai2023gpt4}, Gemini-Pro~\citep{team2023gemini}, \href{https://huggingface.co/HuggingFaceM4/idefics-80b-instruct}{IDEFICS-80B-Instruct} (a reimplementation of Flamingo~\citep{alayrac2022flamingo}), and CogVLM~\citep{wang2023cogvlm}. 
We experiment with two settings:

\paragraph{Image Screenshot + Captions + Accessibility Tree:} This approach provides the accessibility tree representation augmented with image captions as accessibility tree alt-text from BLIP-2-T5XL (similar to the caption-augmented agent), as well as the screenshot of the current webpage as inputs. This provides the model with both the structural information and the visual context of the website.

\paragraph{Image Screenshot + Captions + SoM:} Inspired by Set-of-Marks prompting~\cite{yang2023som}, we perform an initial preprocessing step by using JavaScript to automatically annotate every interactable element on the webpage with a bounding box and a unique ID. The annotated screenshot containing bounding boxes and IDs, are provided as input to the multimodal model along with a text representation of the SoM (see Fig.~\ref{fig:som_figure}). Similar to the baselines above, we also provide the captions from BLIP-2-T5XL for all \texttt{img} elements on the page. There have been several projects\footnote{\href{https://github.com/ddupont808/GPT-4V-Act}{GPT-4V-ACT} and \href{https://github.com/ishan0102/vimGPT}{vimGPT} propose similar interfaces.} that propose similar representations. Most have been proof-of-concept demos, and to the best of our knowledge, we are the first to systematically benchmark this on a realistic and interactive web environment. 


\section{Results and Analysis}

\begin{figure*}[t]
    \centering
    \includegraphics[width=\linewidth]{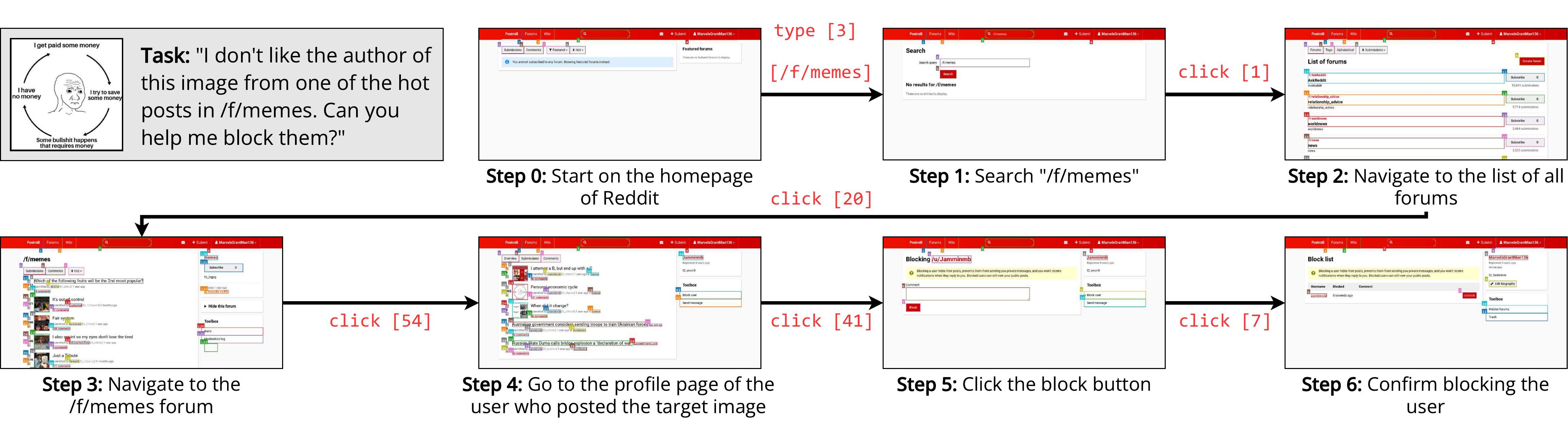}
    \caption{Successful execution trajectory of the GPT-4V + SoM agent on the task for blocking a user that posted a certain picture. The text in red represents the actions output by the agent.} \label{fig:block-user-traj}
\end{figure*}

Our main baseline results are summarized in Tab.~\ref{tab:results}. All existing models substantially underperform compared to humans, which indicate significant headroom in \BenchmarkName{} for future work. We discuss some main findings below with the GPT-4V model, with further analysis in the appendix.

\paragraph{Text-based LLMs Perform Poorly} State-of-the-art text-only LLMs generally achieve poor results, with the best model (GPT-4) achieving an overall success rate of 7.25\%. When we augment the LLMs with captions, this considerably improves success rate (7.25\% to 12.75\% for GPT-4). 

\paragraph{Multimodality Helps} Using multimodal agents significantly improves the success rate: GPT-4V (\texttt{gpt-4-1106-vision-preview}) achieves an overall success rate of 15.05\%, substantially improving over the text-only agents. Gemini-Pro also experiences a significant uplift in success rate, from 3.85\% (caption-augmented) to 6.04\% (multimodal). 
Text-based agents may be limited in their ability to process complex images (e.g., those that require OCR or recognition of non-salient objects). 

\paragraph{SoM Improves Navigability} We observe that the SoM representation (Sec.~\ref{sec:som_method}) further improves the performance of GPT-4V over using the accessibility tree, boosting overall success rate ($15.05\% \rightarrow 16.37\%$). We observe particularly substantial improvements on Classifieds and Reddit, from $12.38\% \rightarrow 17.14\%$ and $8.12\% \rightarrow 9.83\%$ respectively. 
We attribute this to the Classifieds and Reddit websites containing denser visual content. These websites often contain multiple smaller sized images that are arranged very closely (Fig.~\ref{fig:som_figure}). In many of these pages, the accessibility tree does not provide sufficient information to disentangle elements that are spatially close. 
We hypothesize that the SoM representation is superior for strong VLM agents, which can more accurately disentangle and click on the desired elements. 
For the other VLMs, SoM does not significantly improve success, which we attribute to the finding from \citet{yang2023som} that only GPT-4V demonstrates the SoM grounding ability (perhaps due to scale or training data). 

One GPT-4V + SoM execution trajectory that we found particularly compelling was Reddit task \#139, which requires exact image matching to find a post and block a user (Fig.~\ref{fig:block-user-traj}). The model initially attempts to search for the correct forum, and when this fails it navigates to the list of forums. After navigating correctly to \texttt{/f/memes}, it identifies the offending image out of the many images on the page (Step 3 in Fig.~\ref{fig:block-user-traj}) and blocks the author successfully without any unnecessary actions. 

\subsection{Performance by Task Type}  \label{sec:performance_task_subset}

We analyze the success rate of the best VLM agent baseline (GPT-4V + SoM) across several additional subsets of tasks (Tab.~\ref{tab:analysis_subsets}). We include further analysis for other models in Appendix~\ref{sec:appendix_analysis}.

\begin{table}[t] 
\centering
\resizebox{0.85\linewidth}{!}{%
\begin{tabular}{lcc}
  \toprule
  \textbf{Task Subset}\hspace{25mm} & \textbf{\% of Total} & \textbf{SR} ($\uparrow$) \\
  \midrule
  OCR required &  17.1\% &  13.4\% \\
  No OCR required &  82.9\% & \textbf{16.9\%} \\
  \midrule
  Exact image match &  8.7\% &  \textbf{18.9\%} \\
  No exact image match &  91.3\% & 16.2\% \\
  \midrule
  Image inputs &  25.2\% &  \textbf{19.0\%} \\
  No image inputs &  74.8\% & 14.9\% \\
  \bottomrule
\end{tabular}
}
\caption{Success rate (SR) of GPT-4V (SoM) across different types of tasks.}  \label{tab:analysis_subsets}
\end{table}

\paragraph{OCR Tasks} 17.1\% of \BenchmarkName{} require optical character recognition (OCR), such as reading text from product images, or extracting text from an input image. We find that GPT-4V + SoM generally performs worse on tasks that require OCR (13.4\%) compared to tasks which do not (16.9\%), suggesting that OCR may be a bottleneck for current agents.

\paragraph{Exact Image Match} 8.7\% of tasks require exact image matching, which requires agents to identify precise visual matches. GPT-4V + SoM achieves a slightly higher success rate on this subset (18.9\%) compared to other tasks (16.2\%), suggesting that exact image matching is not a primary bottleneck.

\paragraph{Image Input Tasks} 25.2\% of \BenchmarkName{} include one or more input images as part of the objective. These tasks generally appear more tractable for the GPT-4V + SoM agent, and it achieves a higher success rate (19.0\%) compared to tasks without image inputs (14.9\%).

\section{Conclusion}

In this work, we introduced \BenchmarkName{}, a benchmark of realistic tasks designed to rigorously evaluate and advance the capabilities of autonomous multimodal web agents. 
\BenchmarkName{} represents a significant step towards addressing the gap in the evaluation of multimodal agents on visually grounded tasks. 
We also introduce a visual agent inspired by Set-of-Marks prompting, and demonstrate the potential of this approach for simplifying action spaces and improving performance on visually complex websites. 
Our extensive evaluation of state-of-the-art LLM and VLM agents demonstrate that while VLMs show promise, there remains a considerable performance gap compared to humans, who achieve very high success rates on \BenchmarkName{}. 
Our quantitative and qualitative analysis also highlights several common failure modes of existing LLM and VLM agents. We expect future work on improving the reasoning, visual understanding, and planning abilities of agents to be particularly exciting directions.



\section{Ethical and Broader Impacts}

\paragraph{Real World Impacts} Advancing the capabilities of autonomous agents comes with many broader considerations and ethical implications. Strong autonomous agents have the potential to improve the accessibility of computer-based tasks, potentially aiding individuals with disabilities or those lacking technical skills. More broadly, agents have the potential to automate large portions of routine computer work. While the capabilities of existing autonomous agents are insufficient for even simple tasks (as shown in this paper), these impacts highlight the need to ensure that the broader economic and social implications on employment are carefully considered if/when autonomous agents are deployed in real world applications.

\paragraph{Bias and Safety}  When developing autonomous agents, it is also imperative to ensure that these agents do not inadvertently exclude or disadvantage any group. Further analysis is essential to ensure that deployed agents do not exhibit unintended biases. Agents also have the potential to cause more harm (than regular LLMs) in real world applications if careful safeguards are not in place. Further research is necessary to understand and mitigate possible harmful behaviors.

\paragraph{Intended Uses} \BenchmarkName{} is a research benchmark to measure and evaluate the progress of multimodal agents. It is primarily meant to act as a self-contained sandbox environment for safely building robust agents. The models we presented in this paper are research prototypes, and not intended for deployment in practical applications in their current state (especially in high risk domains).

\bibliography{references}

\pagebreak
\appendix
\section*{Appendix}
We provide further information on the collected tasks (Sec~\ref{sec:task_breakdown}), analysis on model failure modes for Gemini and GPT-4 (Sec.~\ref{sec:appendix_analysis}), more details on the new Classifieds environment (Sec.~\ref{sec:appendix_classifieds}), and on the task collection process (Sec.~\ref{sec:appendix_annotation}).

\section{Tasks Breakdown} \label{sec:task_breakdown}

\begin{figure}[h]
\centering
    \includegraphics[width=0.8\linewidth]{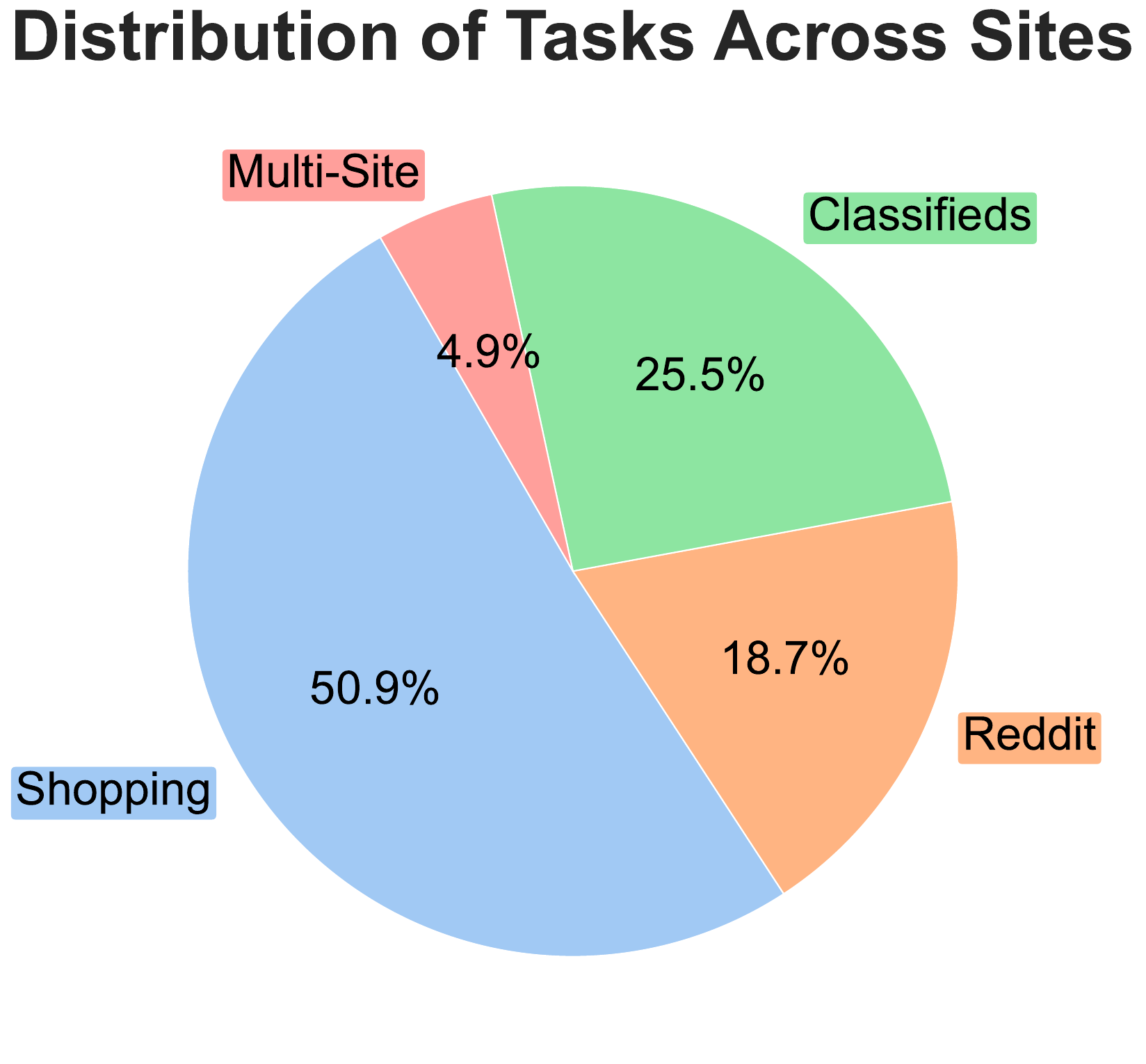}
    \vspace{-0.3in}
    \caption{Tasks proportion by sites.}  
    \label{fig:tasks_breakdown_sites}
\end{figure}

\begin{figure}[h]
\centering
    \includegraphics[width=0.8\linewidth]{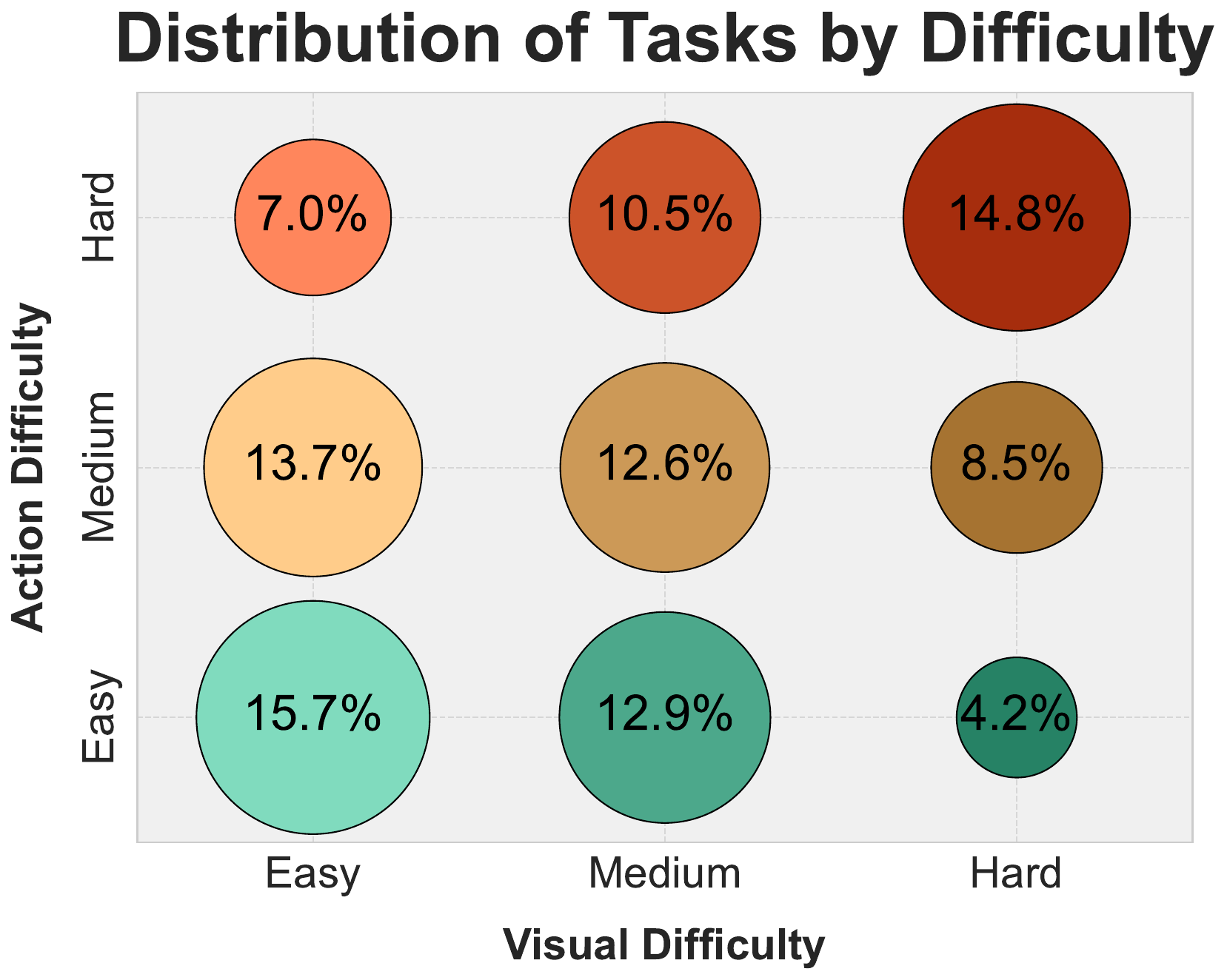}
    \vspace{-0.1in}
    \caption{Tasks proportion by difficulty.}  \label{fig:tasks_breakdown_difficulty}
    \vspace{-0.1in}
\end{figure}

As described in Sec.~\ref{sec:tasks}, we collected a total of 910 tasks across the Classifieds, Reddit, and Shopping sites, with several multi-site tasks that involve more than one site. Several of these tasks also reference Wikipedia as a knowledge base. The breakdown across various sites is summarized in Fig.~\ref{fig:tasks_breakdown_sites}. 

The difficulty level of each task (for both visual difficulty and action difficulty) is summarized in Fig.~\ref{fig:tasks_breakdown_difficulty}, according to the specifications detailed in Sec.~\ref{sec:tasks}. \BenchmarkName{} tasks span a variety of difficulty levels. In Sec.~\ref{sec:performance_difficulty} below, we also discuss the success rate of the agents across difficulty levels, and find that these are roughly correlated, with success rate decreasing as difficulty increases.

\section{Additional Results} \label{sec:additional_reuslts}

\begin{table*}[t]
\centering
\begin{tabular}{@{}llcccc@{}}
\toprule
\multirow{2}{*}{\textbf{Agent Backbone}}  & \multirow{2}{*}{\textbf{Model Type}} & \multicolumn{4}{c}{\textbf{Success Rate ($\uparrow$)}} \\
\cmidrule(lr){3-6}
&  & \textbf{Classifieds} & \textbf{Reddit} & \textbf{Shopping} & \textbf{Overall} \\
\midrule
Llama-3-70B-Instruct &  Caption-augmented & 7.69\% & 5.24\% & 12.88\% &  9.78\% \\ 
Gemini-Flash-1.5 &  Image + Caps + SoM & 3.85\% & 4.76\% & 8.80\% &  6.59\% \\ 
Gemini-Pro-1.5 & Image + Caps + SoM & 5.98\% & 12.86\% & 14.59\% &  11.98\% \\ 
GPT-4o &  Image + Caps + SoM &  20.51\% &  16.67\% & 20.82\% & 19.78\%  \\ 
\bottomrule
\end{tabular}
\caption{Success rates of recent LLM and VLM agents on \BenchmarkName{}.}  \label{tab:additional_results}
\end{table*}

After the ACL submission deadline, we also ran the SoM agent with other recently released frontier VLMs: GPT-4o\footnote{https://openai.com/index/hello-gpt-4o/}, Gemini-Pro 1.5 (\texttt{gemini-1.5-pro-preview-0514}), and Gemini-Flash 1.5 (\texttt{gemini-1.5-flash-preview-0514}). We note that these recent models are \textit{natively multimodal}, which may allow them to achieve stronger performance on multimodal tasks such as VisualWebArena. We also run the Llama-3-70B-Instruct text-only LLM, augmented with captions from BLIP-2. The results are summarized in Tab.~\ref{tab:additional_results}. GPT-4o achieves a success rate of 19.78\%, and outperforms GPT-4V (16.37\%).

Interestingly, we observe that Llama-3-70B-Instruct performs substantially better its Llama-2-70B predecessor, achieving an overall success rate of 9.78\%, which is only slightly below the success rate of the caption-augmented GPT-4 agent (12.75\%), and substantially better than the caption augmented GPT-3.5 (2.97\%) and Llama-2-70B (0.66\%) agents.

\section{Further Analysis}  \label{sec:appendix_analysis}

\subsection{Few-shot Prompting}

\begin{table}[th]
\centering
\vspace{-0.1in}
\resizebox{0.9\linewidth}{!}{%
\begin{tabular}{@{}lcccc@{}}
\toprule
\multirow{2}{*}{\textbf{\# Examples}} & \multicolumn{4}{c}{\textbf{Success Rate ($\uparrow$)}} \\
\cmidrule(lr){2-5}
 & \textbf{Classifieds} & \textbf{Reddit} & \textbf{Shopping} & \textbf{Overall} \\
\midrule
0 & 4.29\% & 2.38\% &  0.43\%  &  2.86\% \\ 
1 & 5.36\% & 1.43\% &  2.14\%  &  3.63\%  \\ 
3 & \textbf{8.15\%} & \textbf{4.29\%} & \textbf{3.42\%} & \textbf{6.04\%} \\ 
\bottomrule
\end{tabular}
}
\caption{Performance with different number of in-context examples.}
\label{tab:few_shot_results}
\end{table}

In most of our main experimental results, we prompt the model with 3 in-context examples. We perform an analysis of the success rate against the number of in-context examples provided (Tab.~\ref{tab:few_shot_results}). For 1-shot experiments, we provide the model with the single in-context example from its corresponding environment. All experiments are run with the multimodal Gemini-Pro model (as GPT-4V is prohibitively expensive) with the Image + Caption + Acc. Tree as the observation space.

We observe that overall success rate tends to increase with the number of examples provided, with a significant jump from 1 to 3 in-context examples. The improved results with a greater number of examples suggest that the performance of the VLM agents may improve significantly if we fine-tune the models on web trajectories, which will be an exciting direction for future work.

\subsection{Performance by Task Difficulty}  \label{sec:performance_difficulty}

\begin{figure*}[ht]
\centering
\resizebox{0.9\linewidth}{!}{%

\begin{tabular}{@{}clrrrrrlrrrr@{}}
\cmidrule(l){2-12}
\multicolumn{1}{l}{}                                              & \multicolumn{11}{c}{Visual Difficulty (v)}                                                                                                                                                                                                                                                                                                                           \\ \cmidrule(l){2-12} 
\multicolumn{1}{|c|}{}                                            & \cellcolor[HTML]{D9D9D9}a \textbackslash v & \cellcolor[HTML]{D9D9D9}Easy   & \cellcolor[HTML]{D9D9D9}Medium & \cellcolor[HTML]{D9D9D9}Hard   & \cellcolor[HTML]{D9D9D9}Overall &  & \cellcolor[HTML]{D9D9D9}a \textbackslash v & \cellcolor[HTML]{D9D9D9}Easy   & \cellcolor[HTML]{D9D9D9}Medium & \cellcolor[HTML]{D9D9D9}Hard   & \cellcolor[HTML]{D9D9D9}Overall \\
\multicolumn{1}{|c|}{}                                         & \cellcolor[HTML]{D9D9D9}Easy               & \cellcolor[HTML]{57BB8A}18.9\%                   & \cellcolor[HTML]{A3DABF}11.1\%                     & \cellcolor[HTML]{A8DCC3}10.5\%                   & \cellcolor[HTML]{7FCCA6}14.8\%                      &  & \cellcolor[HTML]{D9D9D9}Easy               & \cellcolor[HTML]{57BB8A}30.1\%                   & \cellcolor[HTML]{A3DABF}20.5\%                     & \cellcolor[HTML]{75C89F}26.3\%                   & \cellcolor[HTML]{79C9A2}25.8\%                      \\
\multicolumn{1}{|c|}{}                                         & \cellcolor[HTML]{D9D9D9}Medium             & \cellcolor[HTML]{FEFFFF}1.6\%                    & \cellcolor[HTML]{D3EEE1}6.1\%                      & \cellcolor[HTML]{C3E7D5}7.8\%                    & \cellcolor[HTML]{E0F3EA}4.7\%                       &  & \cellcolor[HTML]{D9D9D9}Medium             & \cellcolor[HTML]{CDEBDD}15.2\%                   & \cellcolor[HTML]{ECF8F2}11.3\%                     & \cellcolor[HTML]{E9F7F0}11.7\%                   & \cellcolor[HTML]{DFF3E9}12.9\%                      \\
\multicolumn{1}{|c|}{}                                         & \cellcolor[HTML]{D9D9D9}hard               & \cellcolor[HTML]{FFFFFF}1.6\%                    & \cellcolor[HTML]{E6F5ED}4.2\%                      & \cellcolor[HTML]{FFFFFF}1.5\%                    & \cellcolor[HTML]{F7FCFA}2.4\%                       &  & \cellcolor[HTML]{D9D9D9}hard               & \cellcolor[HTML]{D6EFE3}14.1\%                   & \cellcolor[HTML]{F3FBF7}10.4\%                     & \cellcolor[HTML]{FFFFFF}8.9\%                    & \cellcolor[HTML]{F3FAF7}10.5\%                      \\
\multicolumn{1}{|c|}{}                                         & \cellcolor[HTML]{D9D9D9}Overall            & \cellcolor[HTML]{B7E2CD}9.0\%                    & \cellcolor[HTML]{C7E9D8}7.3\%                      & \cellcolor[HTML]{DFF3E9}4.8\%                    & \cellcolor[HTML]{C8E9D9}7.3\%                       &  & \cellcolor[HTML]{D9D9D9}Overall            & \cellcolor[HTML]{9CD7BA}21.4\%                   & \cellcolor[HTML]{D4EEE1}14.3\%                     & \cellcolor[HTML]{E4F4EC}12.4\%                   & \cellcolor[HTML]{C4E7D6}16.4\%                      \\
\multicolumn{1}{|c|}{}                                            & \multicolumn{5}{c}{(a) Success rate of GPT-4 Text-only}                                                                                                                         &  & \multicolumn{5}{c}{(c) Success rate of GPT-4V + SoM}                                                                                                                            \\
\multicolumn{1}{|c|}{}                                            & \cellcolor[HTML]{D9D9D9}a \textbackslash v & \cellcolor[HTML]{D9D9D9}Easy   & \cellcolor[HTML]{D9D9D9}Medium & \cellcolor[HTML]{D9D9D9}Hard   & \cellcolor[HTML]{D9D9D9}Overall &  & \cellcolor[HTML]{D9D9D9}a \textbackslash v & \cellcolor[HTML]{D9D9D9}Easy   & \cellcolor[HTML]{D9D9D9}Medium & \cellcolor[HTML]{D9D9D9}Hard   & \cellcolor[HTML]{D9D9D9}Overall \\
\multicolumn{1}{|c|}{}                                         & \cellcolor[HTML]{D9D9D9}Easy               & \cellcolor[HTML]{57BB8A}23.1\%                   & \cellcolor[HTML]{80CCA6}18.8\%                     & \cellcolor[HTML]{B5E1CB}13.2\%                   & \cellcolor[HTML]{73C79E}20.1\%                      &  & \cellcolor[HTML]{D9D9D9}Easy               & 6.0                      & 7.7                        & 6.1                      & 6.9                         \\
\multicolumn{1}{|c|}{}                                         & \cellcolor[HTML]{D9D9D9}Medium             & \cellcolor[HTML]{A9DCC3}14.4\%                   & \cellcolor[HTML]{D6EFE3}9.6\%                      & \cellcolor[HTML]{FFFFFF}5.2\%                    & \cellcolor[HTML]{CFECDD}10.4\%                      &  & \cellcolor[HTML]{D9D9D9}Medium             & 10.4                     & 10.6                       & 7.2                      & 10.0                        \\
\multicolumn{1}{|c|}{}                                         & \cellcolor[HTML]{D9D9D9}Hard               & \cellcolor[HTML]{E7F6EE}7.8\%                    & \cellcolor[HTML]{ECF8F2}7.3\%                      & \cellcolor[HTML]{E4F4EC}8.1\%                    & \cellcolor[HTML]{E7F6EE}7.8\%                       &  & \cellcolor[HTML]{D9D9D9}Hard               & 14.1                     & 9.2                        & 12.5                     & 12.1                        \\
\multicolumn{1}{|c|}{}                                         & \cellcolor[HTML]{D9D9D9}Overall            & \cellcolor[HTML]{92D3B3}16.9\%                   & \cellcolor[HTML]{BEE5D2}12.2\%                     & \cellcolor[HTML]{E5F5ED}8.0\%                    & \cellcolor[HTML]{B9E3CE}12.7\%                      &  & \cellcolor[HTML]{D9D9D9}Overall            & 9.5                      & 9.4                        & 10.2                     & 9.6                         \\

\multicolumn{1}{|c|}{\multirow{-12}{*}{\rotatebox[origin=c]{90}{Action Difficulty (a)}}} & \multicolumn{5}{c}{(b) Success rate of GPT-4 + Captions}                                                                                                                        &  & \multicolumn{5}{c}{(d) Trajectory length of GPT-4V + SoM}                                                                                                                      
\end{tabular}

}

\caption{Success rates (a, b, c) and trajectory lengths (d) across different difficulty levels.}  \label{fig:perf_analysis_overall}
\end{figure*}

We conduct an analysis of the GPT-4 models across different action and visual difficulty levels (Fig.~\ref{fig:perf_analysis_overall}). We observe that success rate generally decreases as action/vision difficulty increases, which makes intuitive sense based on the difficulty taxonomy described in Sec.~\ref{sec:tasks}. The findings also show that multimodal models perform better especially on hard visual tasks. On this subset, GPT-4V + SoM achieves an average success rate of 12.4\%, which is significantly higher than that of the caption-augmented (8.0\%) and the text-only agents (4.8\%). In addition to success rates, the GPT-4V trajectory lengths also increased with action difficulty, with harder tasks requiring more steps.

\subsection{Task Subset Analysis}

In this section, we provide more fine-grained analysis across different task subsets, similar to the one in Sec.~\ref{sec:performance_task_subset} of the main paper. We examine both the GPT-4 text and multimodal agents, as well as the Gemini-Pro agents. This analysis may provide useful insights towards capabilities that future VLM models should have to perform well on web navigation tasks (specifically, OCR, exact image matching, and handling multiple interleaved image and text inputs).

\begin{figure*}[h]
    \centering
    \includegraphics[width=0.9\linewidth]{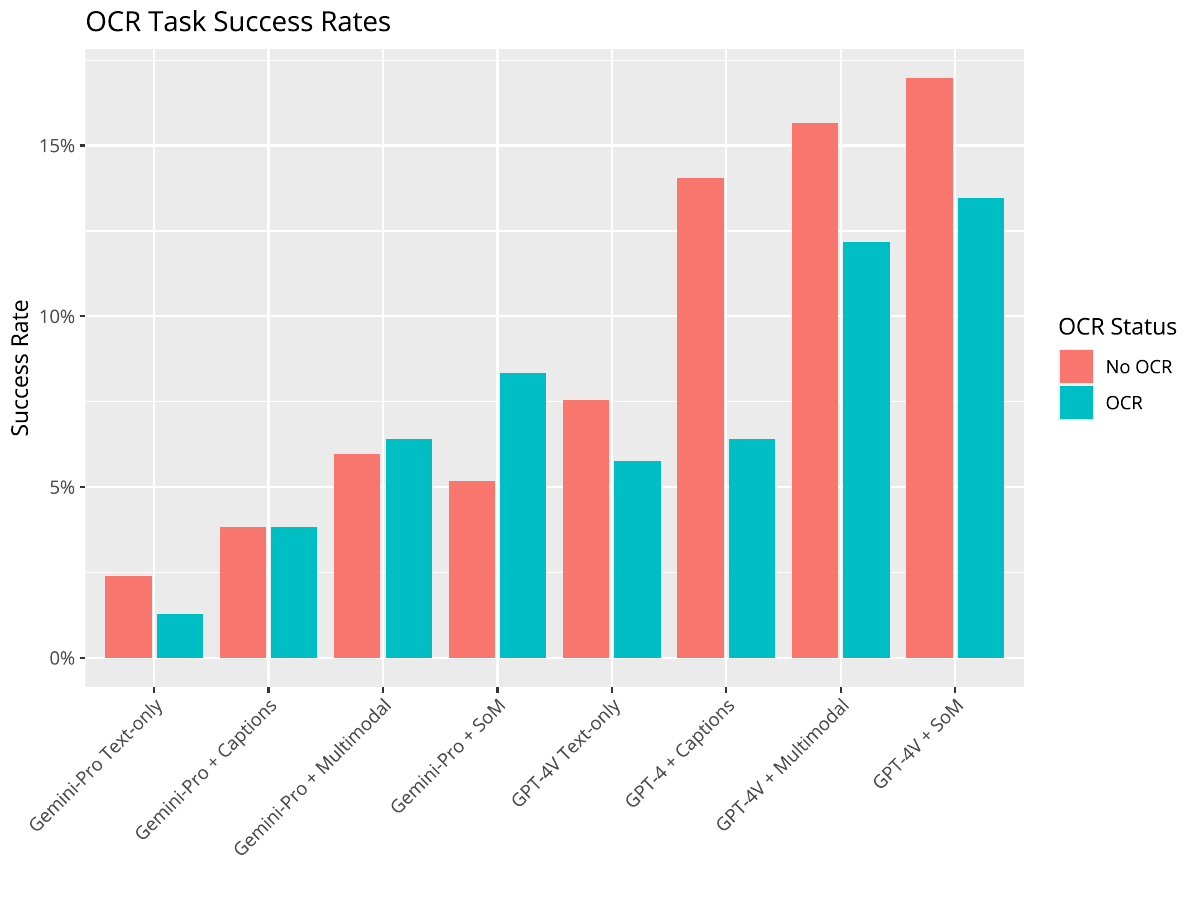}
    \vspace{-0.2in}
    \caption{Success rate of GPT-4 and Gemini agents on tasks that do not require OCR vs. tasks that do.}
    \label{fig:OCR_Success}
\end{figure*}

\paragraph{OCR Tasks} On OCR tasks, which take up 17.1\% of the benchmark, we observe that the GPT-4 family of models achieve a lower success rate on tasks that require OCR compared to tasks that do not (Fig.~\ref{fig:OCR_Success}). This is consistent with the findings for GPT-4V + SoM reported in Sec.~\ref{sec:performance_task_subset} of the main paper. We also observe that introducing multimodality (over just captions) substantially improves performance on OCR tasks (from 6.4\% to 12.2\%), showcasing the importance of having multimodal models for text recognition capabilities, as captioning models generally do not capture such finegrained information.

For Gemini-Pro agents, we also observe similar trends, with the multimodal and SoM models achieving a higher than proportionate gain on the OCR subset (compared to the non-OCR subset). Interestingly, the multimodal Gemini-Pro agents achieve a higher success rate on tasks that require OCR compared to tasks that do not. These results may suggest that it has strong inherent OCR capabilities, which we believe will be useful to explore in future work (especially on the stronger Gemini-Ultra model once it is generally available).

\begin{figure*}[h]
    \centering
    \includegraphics[width=0.9\linewidth]{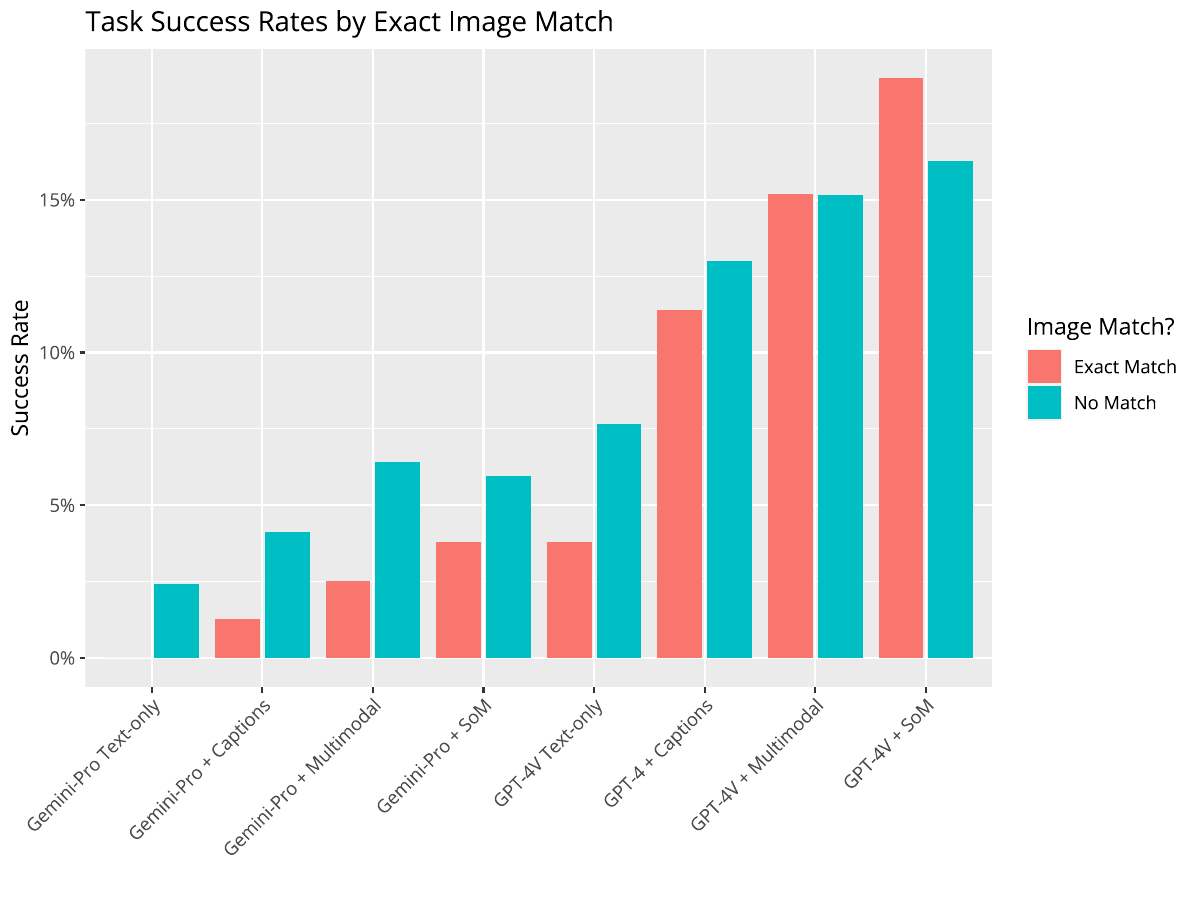}
    \vspace{-0.2in}
    \caption{Success rates of agents on tasks that require exact image match vs. those that do not.}
    \label{fig:Exact_Success}
\end{figure*}

\paragraph{Exact Image Match} Of the tasks in  \BenchmarkName{}, 8.7\% require exact image matching, which tests the ability of agents to identify images that have the exact same content (in contrast to those that are just semantically similar). From Fig.~\ref{fig:Exact_Success}, we 
observe that the GPT-4V SoM model achieves a higher succeess rate on tasks that expect exact image match, while the other GPT-4 agents achieve a relatively lower success rate on the exact match subset. This suggests that the SoM representation may be more optimal for exact image match, due to its visual-centric observation and action space.

For the Gemini models, we observe that success rates on exact match tasks are substantially lower than success rates on non-exact match tasks. Interestingly, we also observe a similar trend as the GPT-4 agents, where introducing multimodality improves success rates on exact match tasks, which is further bolstered with the SoM representation.

\begin{figure*}[h]
    \centering
    \includegraphics[width=0.9\linewidth]{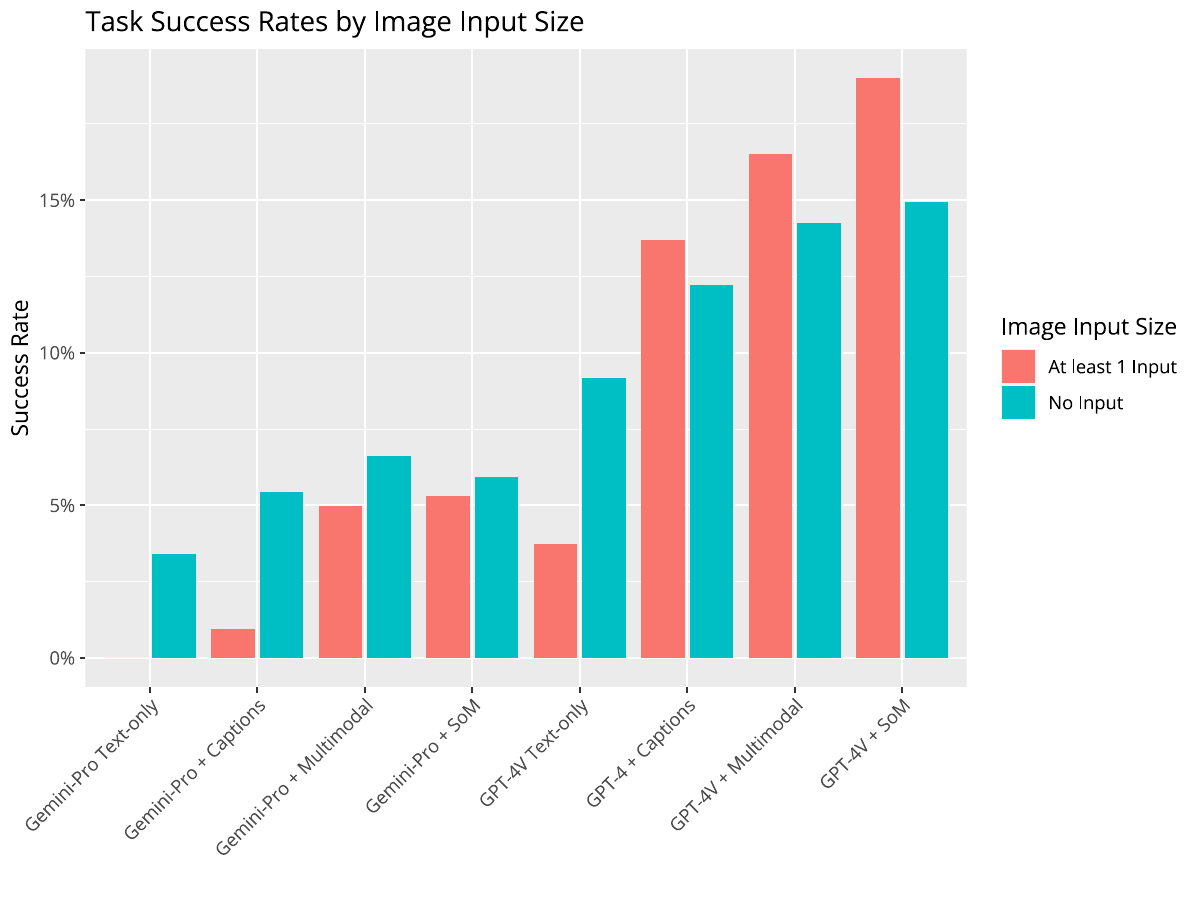}
    \vspace{-0.2in}
    \caption{Success rates of agents on tasks that include input images as part of the specification vs. tasks that are specified with just written text.}
    \label{fig:Input_Success}
\end{figure*}

\paragraph{Image Input Tasks} 25.2\% (229 tasks) in \BenchmarkName{} are specified with image inputs (e.g., the task in Fig.~\ref{fig:block-user-traj}, and the first and third tasks in Fig.~\ref{fig:task_overview}). The results of the Gemini-Pro and GPT-4 agents are summarized in Fig.~\ref{fig:Input_Success}.

We observe that for the GPT-4 agent, success rates are generally higher on tasks that involve image inputs, with the exception of the text-only agent. This aligns with intuition, as agents that do not have access to visual information would not be able to understand the task correctly, and would perform worse at successfully accomplishing it. For the captioning, multimodal, and SoM GPT-4 agents, success rates are higher on the tasks involving image input, which we attribute to these tasks being more tractable once the visual content is correctly understood.

Interestingly, we see a contrast with the Gemini-Pro agents, where success rate is generally lower on tasks that involve input images. This may imply that the model may not be able to process multiple interleaved image-text inputs as well. This may be useful to revisit in the future with the stronger Gemini-Ultra model once it is released, or with stronger open sourced VLMs. We believe that being able to handle interleaved multimodal inputs will be a core requirement for strong web agents, and more comprehensive error analysis with stronger models may yield useful insights.

\begin{figure}[t]
    \centering
    \includegraphics[width=0.9\linewidth]{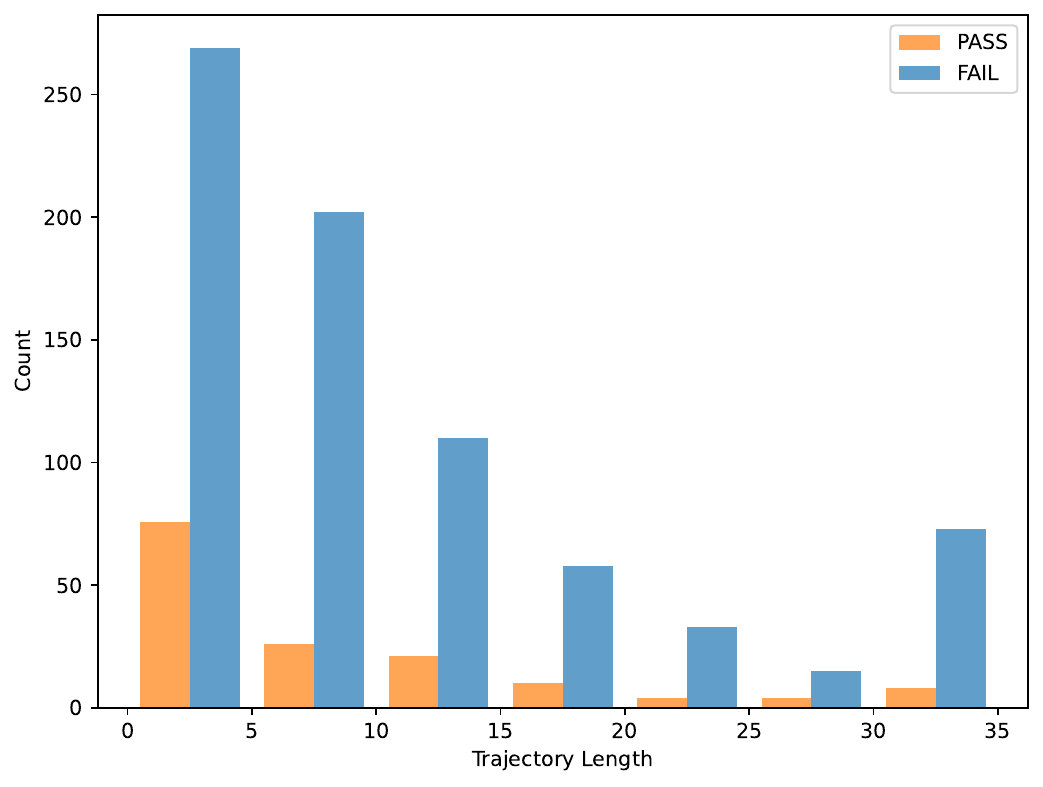}
    \includegraphics[width=0.9\linewidth]{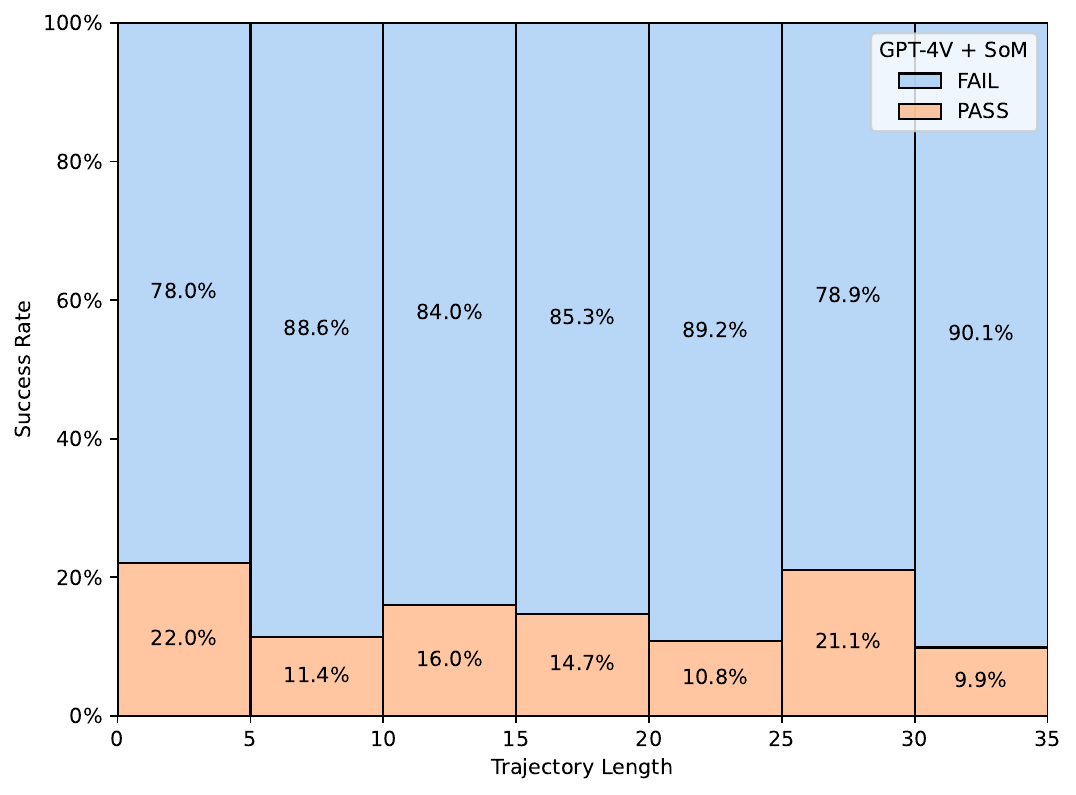}
    \caption{Performance of the GPT-4V + SoM agent across different trajectory lengths.} \label{fig:traj_length_gpt4_som}
\end{figure}

\paragraph{Trajectory Lengths vs. Success Rates}
Hard reasoning tasks, on average, require more steps to be successfully solved. We plot the trajectory length of the GPT-4V + SoM model in Fig.~\ref{fig:traj_length_gpt4_som}. The findings suggest that the model assumes a significant portion of tasks can be completed in a few steps, as it terminates a majority of tasks after less than 10 steps. However, this assumption doesn't imply that the model successfully solves the majority of tasks: the error rate remains relatively uniform across longer trajectory lengths.


\subsection{Failure Modes}

In this section, we describe other common issues we observed with our baseline agent models.

\begin{figure}
    \centering
    \includegraphics[width=1.0\linewidth]{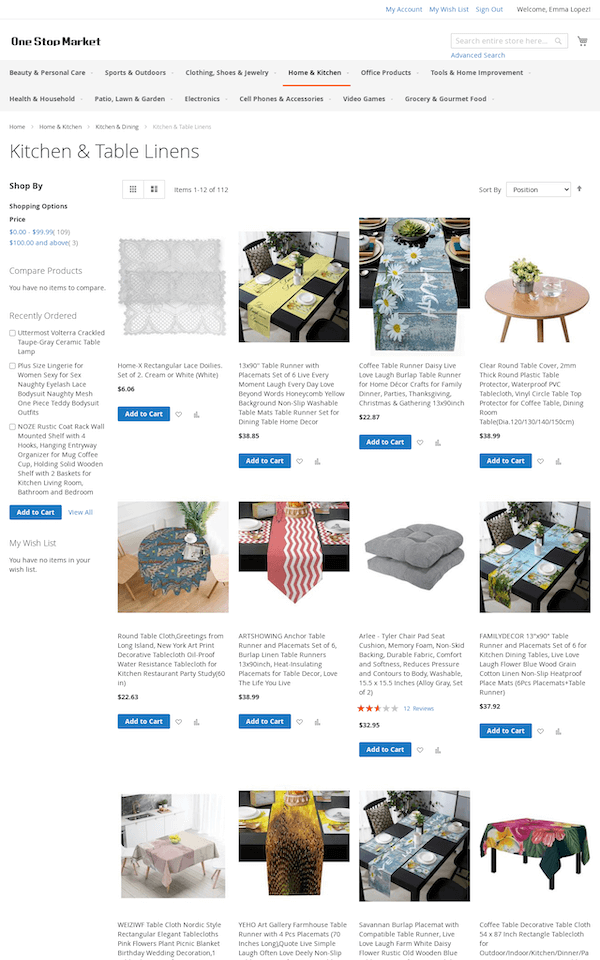}
    \caption{The starting page for the task ``Add the red one in the second row of this page to my shopping cart.''}
    \label{fig:red-tablecloth-example}
\end{figure}

\paragraph{Failure Over Longer Horizons} We observed that in several examples, the agents would correctly perform a task but undo it, leading to failure. 
The GPT-4 captioning-only model on shopping task 54 (``Add the one [poster] with waves to my wish list.'') made an assumption that the product image with a caption about a lighthouse was the correct one, and added it to the wishlist. However, after going to the wish list page the agent removes the poster because ``there is no explicit mention of waves in the current items listed on the Wish List page.'' This issue is not unique to the text input agents; even the GPT-4 SoM agent faced a similar problem in shopping task 397 (``Buy the item on the page with a banana theme.''). The agent initially added the correct item to the shopping cart and proceeded to check out, but stopped in the middle stating in the reasoning trace output that it does not think the item fit the criteria (despite having added it to the cart just a few steps ago).

\paragraph{Failures on Easy Tasks} We observed surprisingly poor performance on many tasks with easy action and easy visual difficulty levels, such as in shopping task 46, which tasks the agent to add the red product in the second row to the cart (starting on the page shown in Fig.~\ref{fig:red-tablecloth-example}). 
The multimodal and SoM GPT-4V agents clicked on a blue tablecloth in the first row and gave up when they couldn't find an option to order it in red. Despite appearing to be a simple task (the correct product is the red cloth in the second row), none of the agents we benchmarked were able to successfully complete it.

\paragraph{Giving Up Too Early} Another frequent issue we observed that occurred across all the agents was giving up too early. For example, GPT-4V + SoM fails on shopping task 248 (``Order a 6 pack of the green chocolate bars. If the shipping is more than 7\% of the total price, leave a 3 star review mentioning it, otherwise 5.''). This tasks involves several steps which the model is able to correctly plan out, but the very first action needed is to slightly scroll down so the ``add to cart'' button is visible. However, even after identifying the correct product the model gives up on the first step instead of scrolling, because it does not immediately see the button. There are other instances of this occurring, such as in shopping task 175, where an agent will use the search bar to search for something, and then immediately give up because it does not see the target product instead of trying new ways to find the product.

\paragraph{Getting Stuck in Loops} Another issue we observed was oscillating or looping between pages, where the agent would look something up or navigate to a page, unsuccessfully attempt to perform the next action (such as adding it to the cart), and on failure it goes back and repeats from the beginning. An example of this is in classifieds task 205 where the model is tasked to compare two makeup palettes in two tabs, and the GPT-4V agent spends most of the time switching between the tabs. We believe that these issues will likely be alleviated by introducing more sophisticated tracking of past states and execution history, which is a promising direction for future work.

\paragraph{Failure Example: Changing User Phone Number}
\begin{figure*}[t]
    \centering
    \includegraphics[width=\linewidth]{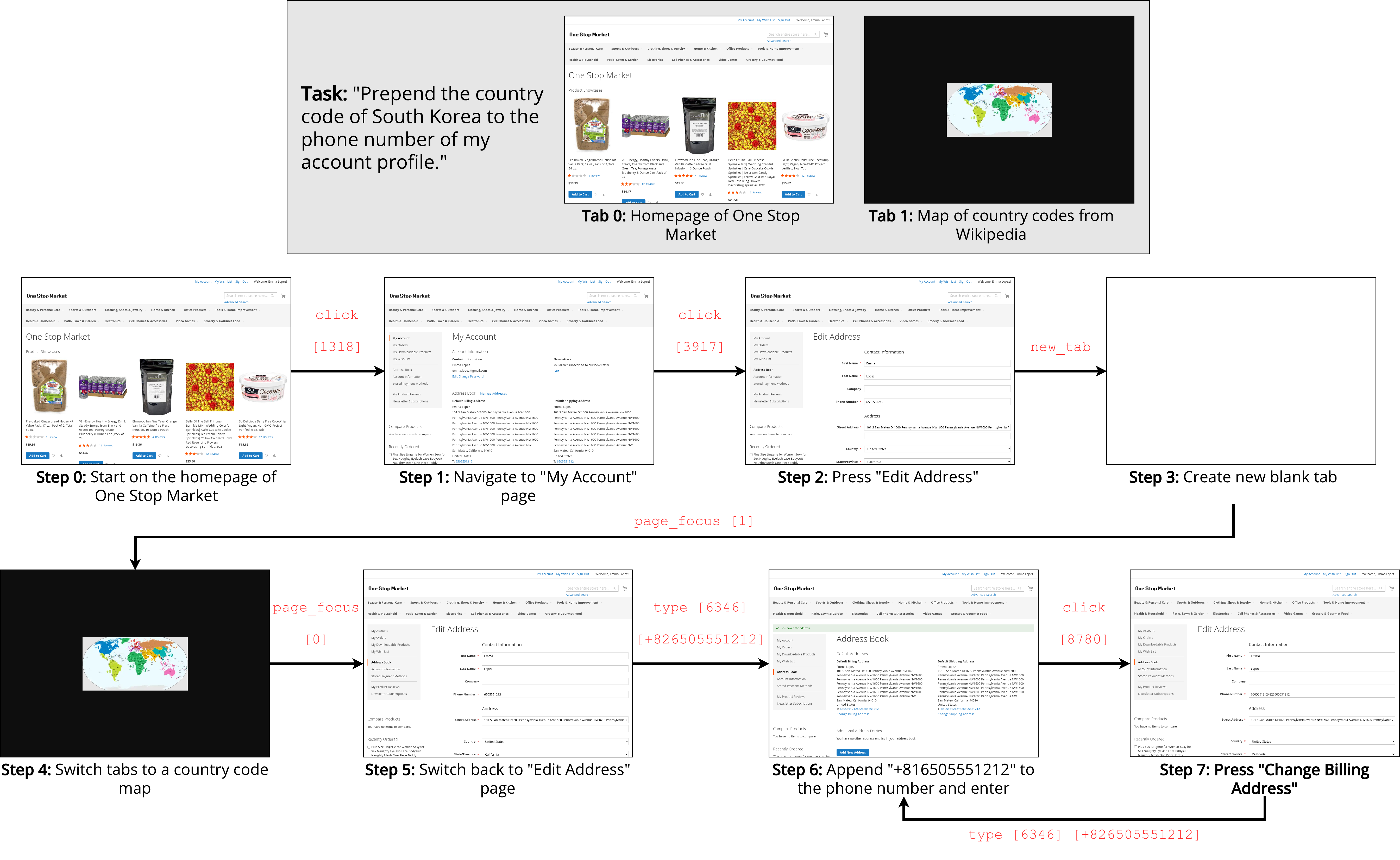}
    \vspace{-0.25in}
    \caption{Unsuccessful execution trajectory of the GPT-4V multimodal agent on the task for adding the a country code to the user's phone number. The text in red represents the commands output by the agent.}
    \label{fig:country-code-traj}
\end{figure*}

Shopping task \#345, a multi-site task that also involves the Wikipedia site, demonstrated several interesting points of failure that we saw throughout the execution traces for many other tasks. Fig.~\ref{fig:country-code-traj} contains the execution trace of the GPT-4V multimodal agent for the task ``Prepend the country code of South Korea to the phone number of my account profile.'' There are three major mistakes made by the agent in this execution trace:

\begin{itemize}
    \item \textbf{Useless actions:} In step 3 of the trajectory, the agent creates a new blank tab and does not interact with it for the rest of the trajectory. While this does not impact the correctness of the final task, it does show that the agents sometimes take unnecessary steps.

    \item \textbf{Appending text instead of replacing:} Many agents added text to input fields without deleting the previous text, which would often result in long, repeating search queries or addresses. An example of this occurs in step 7 of Fig.~\ref{fig:country-code-traj}.

    \item \textbf{Repeating actions:} Another frequent issue we saw across agents was repeating actions, like how the agent kept jumping between step 6 and step 7 of Fig.~\ref{fig:country-code-traj} until it hit the maximum trajectory length. In this case, we believe this looping effect stems from the issue mentioned above and each time the agent tries to correct the phone number, it keeps appending the correct number instead of replacing the incorrect number with the correct number.
    
\end{itemize}

\subsection{Comparison Between Agents}

In this section, we describe some qualitative differences we observed between the different agents on various tasks in \BenchmarkName{}.

\paragraph{Text-only vs. Caption-augmented Agents} For GPT-4, the text model unsurprisingly performs much worse than even the captioning model, failing to even do the most basic tasks. For example, Reddit task \#101 is the relatively simple task to ``Navigate to the comments section of a post that contains a picture of a keyboard.'' Out of all of the GPT-4 baseline agents, the text-only agent is the only one to fail this task, as it's unable to identify the appropriate post from just the title. Interestingly, it still manages to make an educated guess and navigate to the hottest post on /f/MechanicalKeyboards (which unfortunately, did not include a keyboard in its image).

\paragraph{Caption-augmented vs. SoM Agents} We observed in many examples that the GPT-4V SoM and multimodal agents outperformed the caption-augmented baselines in terms of navigation capabilities and visual understanding. The multimodal models were generally better at understanding visual information on webpages, as relevant information in many images is lost when they are translated into captions. One pertinent example is Reddit task \#40, where a picture of the skyline of Pittsburgh is provided, and the task is ``I'd like to find the subreddit for the city this photo was taken in. Can you navigate to it?''. The GPT-4V + SoM agent correctly identifies the location of the photo, with the first line of its reasoning output as \textit{``The photo shows a city skyline with prominent buildings labeled with logos for UPMC and PNC, which suggests that the photo was taken in Pittsburgh, Pennsylvania.''}. Using this information, the agent is able to successfully navigate to the appropriate subreddit, /f/pittsburgh. In contrast, the captioning agent labels the image as ``city skyline with many tall buildings'' (as this is the output from the BLIP-2 model), which prevents it from identifying the appropriate subreddit. This highlights a fundamental issue with captioning models: they frequently highlight salient visual information, which hinders success on many tasks that require fine-grained visual information not typically captured by captions.

\begin{figure*}[t]
    \centering
    \includegraphics[width=0.95\linewidth]{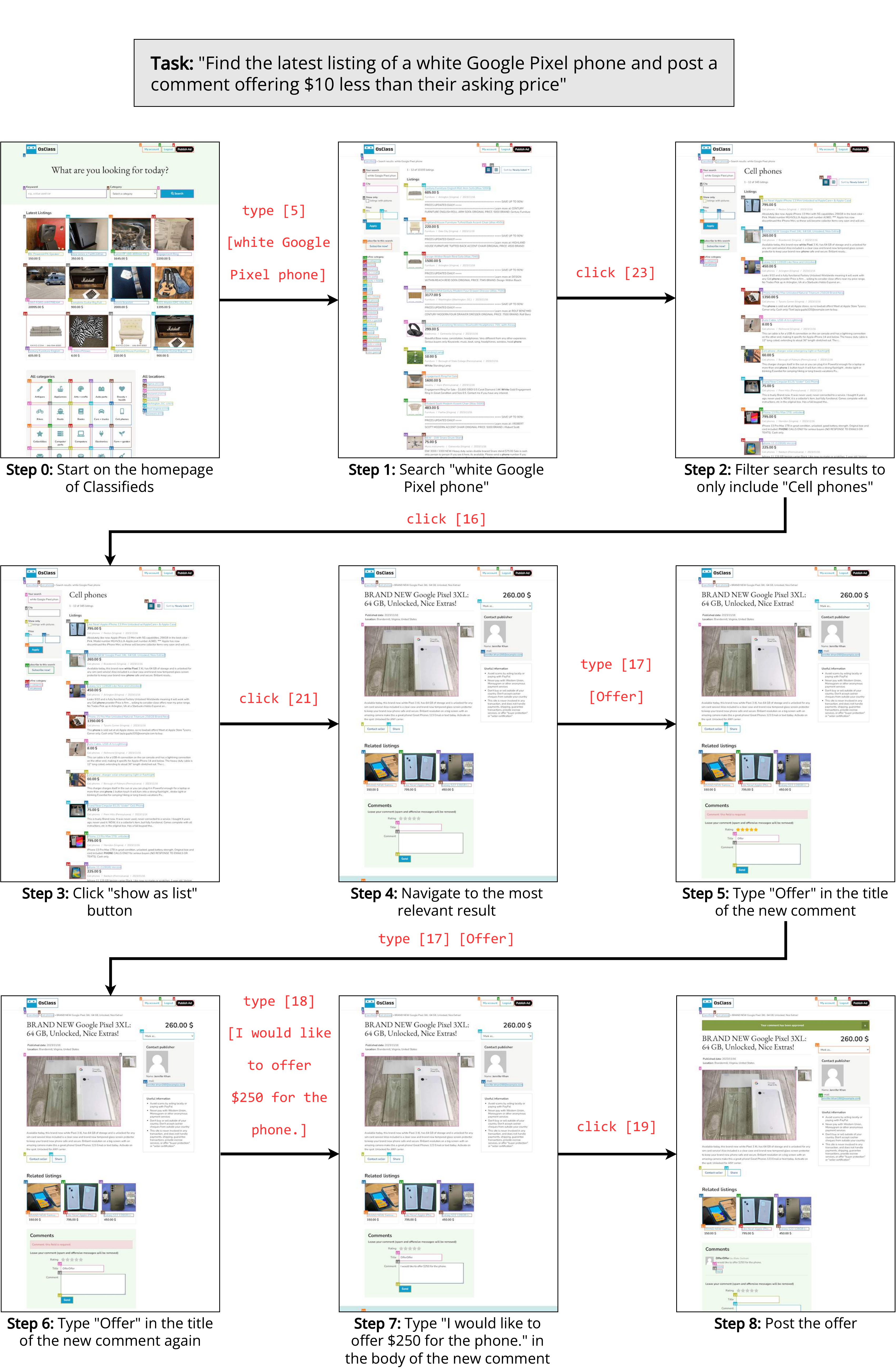}
    \caption{GPT-4V SoM agent on Classifieds task \#31, ``Find the latest listing of a white Google Pixel phone and post a comment offering \$10 less than their asking price.''. It succeeds at the task by leveraging the more efficient navigation space.}
    \label{fig:classifieds_task31}
\end{figure*}

\paragraph{Multimodal vs. SoM Agents} The SoM representation generally performs better on tasks that require more navigation steps, due to its simplified observation and action space. One example is classifieds task \#31, ``Find the latest listing of a white Google Pixel phone and post a comment offering \$10 less than their asking price.'' (Fig.~\ref{fig:classifieds_task31}). While the multimodal model was unable to search for the correct terms, the SoM model was able to leverage the simplified action space to traverse more efficiently throughout the environment. It succeeded at this task by filtering for cell phones after the initial search for more relevant results, and managed to fill out the necessary comment form fields. 
From our observations, the SoM representation is generally more efficient compared to the multimodal representation (which only has access to the page screenshot and accessibility tree). With a strong VLM capable of SoM, the agent does not have to implicitly perform visual co-referencing to match elements from the accessibility tree to the visual buttons and inputs that it wants to interact with.

\section{The Classifieds Environment}  \label{sec:appendix_classifieds}
\begin{figure*}[ht]
\centering
  \includegraphics[width=\textwidth]{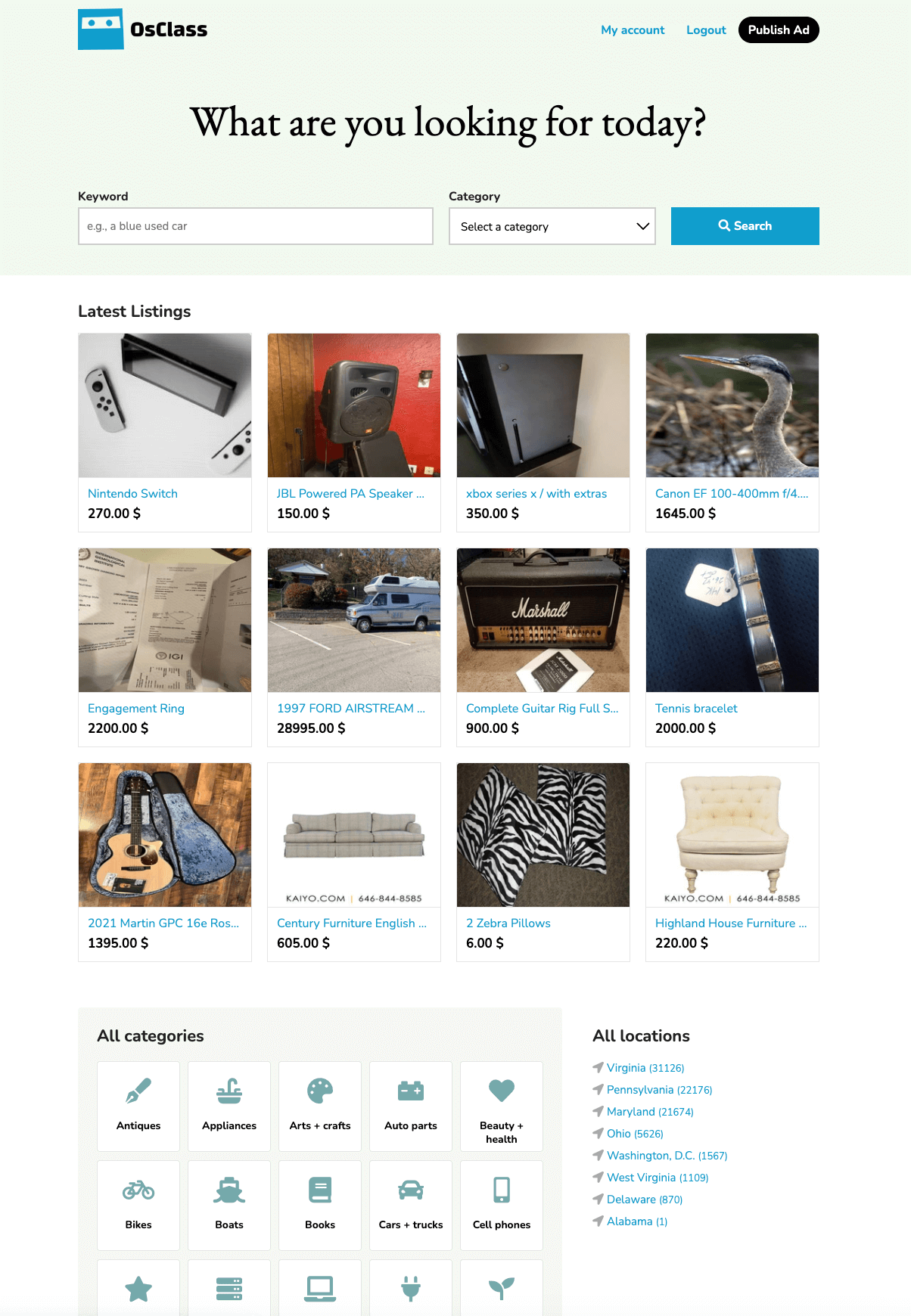}
  \caption{Homepage of the Classifieds site. Users can search for keywords, filter by category, or post location.}
  \label{fig:classifieds_homepage}
\end{figure*}

\begin{figure*}[ht]
  \centering
  \includegraphics[width=\textwidth]{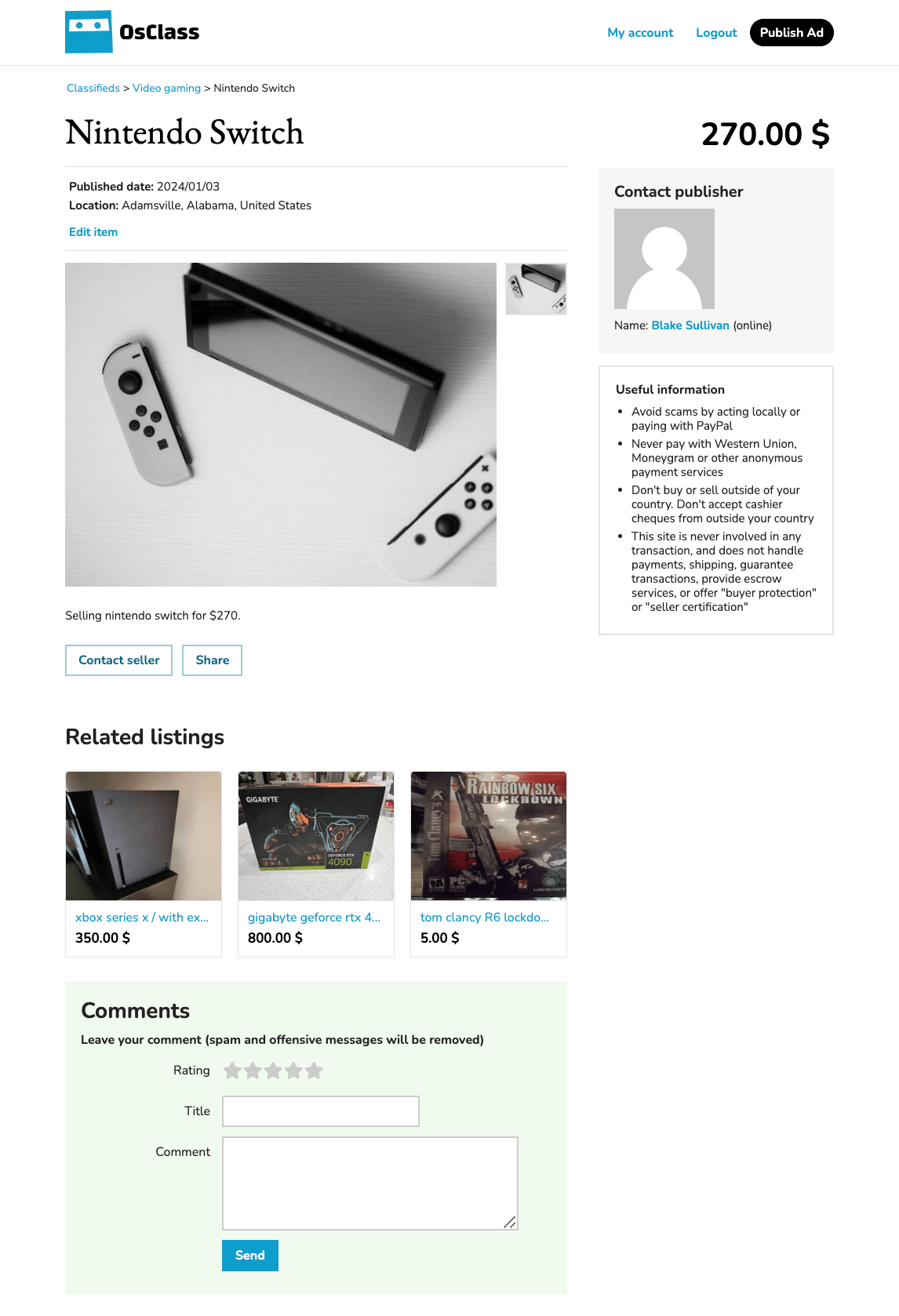}
  \caption{Example post in the Classifieds website. Users can add comments and reviews to individual listings.}
  \label{fig:classifieds_detail}
\end{figure*}

The Classifieds environment contains 65,955 listings, each with a title, text description, and a product image of the item being sold. To populate the site with realistic content, we scraped data across a variety of categories on Craigslist over 3 weeks, focusing on the Northeastern States of the US (similar to the geographic region in the Reddit site). This approach ensured a diverse and rich dataset, representative of real-world classifieds posts. 
We utilized the \texttt{scrubadub} Python package for redacting Personally Identifiable Information (PII), including addresses, phone numbers, and emails. We use generated placeholders for names (e.g., ``Bill Smith''), emails with fictitious addresses (e.g., \texttt{bill\_smith@example.com}), and phone numbers with the fictional 555-prefix numbers.

Fig.~\ref{fig:classifieds_homepage} and \ref{fig:classifieds_detail} show two pages within the Classifieds site, the homepage and the detail page of a particular listing. Users can also use the search function, or filter posts by category or location to find items.

\section{Task Collection Process}  \label{sec:appendix_annotation}
Our main task collection process is described in Sec.~\ref{sec:tasks}. We collected the set of 910 tasks by recruiting 6 computer science graduate students (co-authors of this paper), who were all familiar with commercial versions of the Classifieds, Shopping, and Reddit sites, and have used them in their personal lives.

Annotators were first instructed to spend some time exploring the \BenchmarkName{} websites, to familiarize themselves with their functionality and content (as this may differ slightly from real world implementations). During task creation, we encouraged annotators to be creative, and make use of the visual layouts of the websites, input images, and cross-site functionalities to develop creative and realistic tasks. We ensured that there were no repeated tasks, and that there were not too many tasks of the same type (by first producing templates, followed by instantiating them with different arguments to create multiple tasks, as described in Sec.~\ref{sec:tasks}).

\section{Baseline Agent Settings}

\begin{figure*}[th]
\noindent\fbox{\parbox{\textwidth}{%
\small
You are an autonomous intelligent agent tasked with navigating a web browser. You will be given web-based tasks. These tasks will be accomplished through the use of specific actions you can issue.
\\~\\
Here's the information you'll have: \\
The user's objective: This is the task you're trying to complete.\\
The current web page screenshot: This is a screenshot of the webpage, with each interactable element assigned a unique numerical id. Each bounding box and its respective id shares the same color.\\
The observation, which lists the IDs of all interactable elements on the current web page with their text content if any, in the format [id] [tagType] [text content]. tagType is the type of the element, such as button, link, or textbox. text content is the text content of the element. For example, [1234] [button] ['Add to Cart'] means that there is a button with id 1234 and text content 'Add to Cart' on the current web page. [] [StaticText] [text] means that the element is of some text that is not interactable.\\
The current web page's URL: This is the page you're currently navigating.\\
The open tabs: These are the tabs you have open.\\
The previous action: This is the action you just performed. It may be helpful to track your progress.
\\~\\
The actions you can perform fall into several categories:
\\~\\
Page Operation Actions:\\
\texttt{\`}\texttt{\`}\texttt{\`}click [id]\texttt{\`}\texttt{\`}\texttt{\`}: This action clicks on an element with a specific id on the webpage.\\
\texttt{\`}\texttt{\`}\texttt{\`}type [id] [content]\texttt{\`}\texttt{\`}\texttt{\`}: Use this to type the content into the field with id. By default, the ``Enter'' key is pressed after typing unless press\_enter\_after is set to 0, i.e., \texttt{\`}\texttt{\`}\texttt{\`}type [id] [content] [0]\texttt{\`}\texttt{\`}\texttt{\`}.\\
\texttt{\`}\texttt{\`}\texttt{\`}hover [id]\texttt{\`}\texttt{\`}\texttt{\`}: Hover over an element with id.\\
\texttt{\`}\texttt{\`}\texttt{\`}press [key\_comb]\texttt{\`}\texttt{\`}\texttt{\`}:  Simulates the pressing of a key combination on the keyboard (e.g., Ctrl+v).\\
\texttt{\`}\texttt{\`}\texttt{\`}scroll [down]\texttt{\`}\texttt{\`}\texttt{\`} or \texttt{\`}\texttt{\`}\texttt{\`}scroll [up]\texttt{\`}\texttt{\`}\texttt{\`}: Scroll the page up or down.
\\~\\
Tab Management Actions:\\
\texttt{\`}\texttt{\`}\texttt{\`}new\_tab\texttt{\`}\texttt{\`}\texttt{\`}: Open a new, empty browser tab.\\
\texttt{\`}\texttt{\`}\texttt{\`}tab\_focus [tab\_index]\texttt{\`}\texttt{\`}\texttt{\`}: Switch the browser's focus to a specific tab using its index.\\
\texttt{\`}\texttt{\`}\texttt{\`}close\_tab\texttt{\`}\texttt{\`}\texttt{\`}: Close the currently active tab.
\\~\\
URL Navigation Actions:\\
\texttt{\`}\texttt{\`}\texttt{\`}goto [url]\texttt{\`}\texttt{\`}\texttt{\`}: Navigate to a specific URL.\\
\texttt{\`}\texttt{\`}\texttt{\`}go\_back\texttt{\`}\texttt{\`}\texttt{\`}: Navigate to the previously viewed page.\\
\texttt{\`}\texttt{\`}\texttt{\`}go\_forward\texttt{\`}\texttt{\`}\texttt{\`}: Navigate to the next page (if a previous 'go\_back' action was performed).
\\~\\
Completion Action:\\
\texttt{\`}\texttt{\`}\texttt{\`}stop [answer]\texttt{\`}\texttt{\`}\texttt{\`}: Issue this action when you believe the task is complete. If the objective is to find a text-based answer, provide the answer in the bracket.
\\~\\
Homepage:\\
If you want to visit other websites, check out the homepage at http://homepage.com. It has a list of websites you can visit.\\
http://homepage.com/password.html lists all the account name and password for the websites. You can use them to log in to the websites.
\\~\\
To be successful, it is very important to follow the following rules:\\
1. You should only issue an action that is valid given the current observation\\
2. You should only issue one action at a time.\\
3. You should follow the examples to reason step by step and then issue the next action.\\
4. Generate the action in the correct format. Start with a ``In summary, the next action I will perform is'' phrase, followed by action inside \texttt{\`}\texttt{\`}\texttt{\`}\texttt{\`}\texttt{\`}\texttt{\`}. For example, ``In summary, the next action I will perform is \texttt{\`}\texttt{\`}\texttt{\`}click [1234]\texttt{\`}\texttt{\`}\texttt{\`}''.\\
5. Issue stop action when you think you have achieved the objective. Don't generate anything after stop.
}
}
\caption{System message used in our SoM agent.}  \label{fig:system_message}
\end{figure*}

\begin{figure*}[th]
\noindent\fbox{\parbox{\textwidth}{%
\small
{\color{red}{example\_user:}} \\
IMAGES: (1) current page screenshot\\
OBSERVATION:\\
\lbrack31\rbrack\ [IMG]\ \lbrack Image, description: hp fx-7010dn fax machine, url: B08GKZ3ZKD.0.jpg\rbrack\\
\lbrack32\rbrack\ [A]\ \lbrack HP CB782A\#ABA 640 Inkjet Fax Machine (Renewed)\rbrack\\
\lbrack\ \rbrack\ [StaticText]\ \lbrack\$279.49\rbrack\\
\lbrack33\rbrack\ [BUTTON]\ \lbrack Add to Cart\rbrack\\
\lbrack34\rbrack\ [A]\ \lbrack Add to Wish List\rbrack\\
\lbrack35\rbrack\ [A]\ \lbrack Add to Compare\rbrack\\
URL: http://onestopmarket.com/office-products/office-electronics.html\\
OBJECTIVE: What is the price of HP Inkjet Fax Machine?\\
PREVIOUS ACTION: None
\\
{\color{red}{example\_assistant:}} Let's think step-by-step. This page list the information of HP Inkjet Fax Machine, which is the product identified in the objective. Its price is \$279.49. I think I have achieved the objective. I will issue the stop action with the answer. In summary, the next action I will perform is \texttt{\'}\texttt{\'}\texttt{\'}stop [\$279.49]\texttt{\'}\texttt{\'}\texttt{\'}
\\~\\
{\color{red}{example\_user:}} \\
IMAGES: (1) current page screenshot\\
OBSERVATION:\\
\lbrack\ \rbrack\ [StaticText]\ \lbrack/f/food\rbrack\\
\lbrack\ \rbrack\ [StaticText]\ \lbrack[homemade] Obligatory Halloween Pumpkin Loaf!\ \ \ Submitted by\ \ kneechalice\ \ t3\_yid9lu\ \ 1 year ago\rbrack\\
\lbrack9\rbrack\ [IMG]\ \lbrack\ \rbrack\\
\lbrack\ \rbrack\ [StaticText]\ \lbrack Submitted by\ \ kneechalice\ \ t3\_yid9lu\ \ 1 year ago\rbrack\\
\lbrack10\rbrack\ [A]\ \lbrack kneechalice\rbrack\\
\lbrack11\rbrack\ [A]\ \lbrack 45 comments\rbrack\\
\lbrack\ \rbrack\ [StaticText]\ \lbrack[I ate] Maple Pecan Croissant\ \ \ Submitted by\ \ AccordingtoJP\ \ t3\_y3hrpn\ \ 1 year ago\rbrack\\
\lbrack14\rbrack\ [IMG]\ \lbrack\ \rbrack\\
\lbrack\ \rbrack\ [StaticText]\ \lbrack Submitted by\ \ AccordingtoJP\ \ t3\_y3hrpn\ \ 1 year ago\rbrack\\
\lbrack15\rbrack\ [A]\ \lbrack AccordingtoJP\rbrack\\
\lbrack16\rbrack\ [A]\ \lbrack 204 comments\rbrack\\
URL: http://reddit.com\\
OBJECTIVE: Tell me what the top comment on the croissant post says.\\
PREVIOUS ACTION: None
\\
{\color{red}{example\_assistant:}} Let's think step-by-step. This page has a post titled '\lbrack I ate\rbrack \ Maple Pecan Croissant', which is the post mentioned in the objective. In order to find the top comment, I will navigate into the comments section of the post. In summary, the next action I will perform is \texttt{\'}\texttt{\'}\texttt{\'}click \lbrack11\rbrack\texttt{\'}\texttt{\'}\texttt{\'}
\\~\\
{\color{red}{example\_user:}} \\
IMAGES: (1) current page screenshot\\
OBSERVATION:\\
\lbrack\ \rbrack\ [StaticText]\ \lbrack What are you looking for today?\rbrack\\
\lbrack5\rbrack\ [INPUT]\ \lbrack\ \rbrack\\
\lbrack6\rbrack\ [SELECT]\ \lbrack Select a category\rbrack\\
\lbrack7\rbrack\ [BUTTON]\ \lbrack Search\rbrack\\
\lbrack\ \rbrack\ [StaticText]\ \lbrack Latest Listings\rbrack\\
\lbrack\ \rbrack\ [StaticText]\ \lbrack Atlas Powered Audio System w/ Tripod\ \ 150.00 \$\ \ Music instruments\ \ Borough of Red Lion\ \ (Pennsylvania)\ \ 2023/11/16\rbrack\\
\lbrack8\rbrack\ [IMG]\ \lbrack Atlas Powered Audio System w/ Tripod\rbrack\\
\lbrack9\rbrack\ [A]\ \lbrack Atlas Powered Audio System w/ Tripod\rbrack\\
\lbrack\ \rbrack\ [StaticText]\ \lbrack 150.00 \$\rbrack\\
\lbrack\ \rbrack\ [StaticText]\ \lbrack Neptune Gaming Console\ \ 350.00 \$\ \ Video gaming\ \ Pennwyn\ \ (Pennsylvania)\ \ 2023/11/16\rbrack\\
\lbrack10\rbrack\ [IMG]\ \lbrack Neptune Gaming Console\rbrack\\
\lbrack11\rbrack\ [A]\ \lbrack Neptune Gaming Console\rbrack\\
\lbrack\ \rbrack\ [StaticText]\ \lbrack 350.00 \$\rbrack\\
URL: http://classifieds.com\\
OBJECTIVE: Help me find the cheapest dark colored guitar.\\
PREVIOUS ACTION: None
\\
{\color{red}{example\_assistant:}} Let's think step-by-step. The objective is to find the cheapest dark colored guitar on the site. The site has a search box whose ID is \lbrack5\rbrack. I can search for guitars by entering ``guitar''. I can submit this by pressing the Enter afterwards. In summary, the next action I will perform is \texttt{\'}\texttt{\'}\texttt{\'}type \lbrack5\rbrack\ \lbrack guitar\rbrack\  \lbrack1\rbrack\texttt{\'}\texttt{\'}\texttt{\'}
}
}
\caption{In-context examples and prompt used for our SoM agents. The webpage screenshots with SoM are added to the end of each round of the {\color{red}{example\_user}} dialogue.}  \label{fig:prompt}
\end{figure*}

For all baseline agents we report in the paper, we use a webpage viewport size of $1280 \times 2048$, and truncate text observations to 3840 tokens (or 15360 characters for Gemini). For models with shorter context windows (e.g., LLaMA, IDEFICS, CogVLM), we instead use a viewport size of $1280 \times 720$ and truncate text observations to 640 tokens. For GPT-3.5 and GPT-4 models, we follow \cite{zhou2023webarena} in using a temperature of 1.0 and a top-p of 0.9. For Gemini models we use the suggested default temperature of 0.9 and top-p of 1.0. For the remaining models, we find that they benefit from sampling from lower temperatures, and use a temperature of 0.6 and top-p of 0.95. Nucleus sampling~\citep{holtzman2019curious} is used in all experiments. 

The system message and the prompt with in-context examples for the baseline SoM agents are shown in Fig.~\ref{fig:system_message} and Fig.~\ref{fig:prompt} respectively. We prompt the model with 3 in-context examples for all baselines. For multimodal and SoM models, we include the screenshot of each in-context example as well as the screenshot of the current page. For text-only and caption augmented models, the examples consist of just the text and captions.

\end{document}